%% file: main.tex
\documentclass[10pt,letterpaper]{article}
\usepackage[top=0.85in,left=2.75in,footskip=0.75in]{geometry}

\usepackage{amsmath,amssymb}

\usepackage{changepage}

\usepackage[utf8x]{inputenc}

\usepackage{textcomp,marvosym}

\usepackage{cite}

\usepackage{nameref,hyperref}

\usepackage[right]{lineno}

\usepackage{microtype}
\DisableLigatures[f]{encoding = *, family = * }

\usepackage[table]{xcolor}

\usepackage{array}

\newcolumntype{+}{!{\vrule width 2pt}}

\newlength\savedwidth

\raggedright
\usepackage[aboveskip=1pt,labelfont=bf,labelsep=period,justification=raggedright,singlelinecheck=off]{caption}

\makeatletter
\renewcommand{\@biblabel}[1]{\quad#1.}
\makeatother

\usepackage{lastpage,fancyhdr,graphicx}
\usepackage{epstopdf}

\usepackage{bm}
\newcommand\R{\mathrm}
\newcommand\B{\mathbf}
\usepackage{siunitx}
\usepackage{upgreek}

\newcommand{\beginsupplement}{%
        \setcounter{table}{0}
        \renewcommand{\thetable}{S\arabic{table}}%
        \setcounter{figure}{0}
        \renewcommand{\thefigure}{S\arabic{figure}}%
     }

\fancyheadoffset[L]{2.25in}
\fancyfootoffset[L]{2.25in}
\begin{document}
\vspace*{0.2in}

\begin{flushleft}
{\Large
\textbf\newline{Bayesian Inference in High-Dimensional Time-Series with the Orthogonal Stochastic Linear Mixing Model} 
}
\newline
\\
Rui Meng\textsuperscript{1},
Kristofer Bouchard\textsuperscript{1},
\\
\bigskip
\textbf{1} Lawrence Berkeley National Laboratory, Berkeley, CA, US
\\
\bigskip

%
%





* rmeng@lbl.gov

\end{flushleft}
\input{ABS.tex}


\vspace{-10pt}
\input{INTRO.tex}

\vspace{-10pt}
\input{METHOD.tex}

\vspace{-10pt}
\input{RESULT.tex}

\vspace{-10pt}
\input{DIS.tex}

\section*{Supporting information}
\beginsupplement
\input{SUP.tex}

\clearpage
\small
\bibliography{MyBib.bib}

\end{document}

%% file: ABS.tex
\section*{Abstract}
Understanding the structure of neural activities from multiple single-trial measurements plays an important role in neuroscience, which directly contributes to a deeper understanding of brain function as well as behavior. However, extracting useful information from the multiple single-trial measurements is hard due to two challenges. First, the number of simultaneously recorded neurons is large. Second, the dynamics of neural population activity is complex. Gaussian-process factor analysis (GPFA) is one of most popular methods to capture structure across multiple neurons and extract useful latent trajectories. However, this class of models typically assumes that the correlations across neurons are time-invariant. We develop the Stochastic linear mixing models (SLMM) by assuming the mixture coefficients depend on input (time), making them more flexible and effective for capturing complex neural dynamics. However, the inference for SLMMs is currently intractable for large datasets, making them difficult to use on modern neural timeseries datasets. Therefore, in this work, we propose a new regression framework, the orthogonal stochastic linear mixing model (OSLMM) that introduces an orthogonal constraint amongst the mixing coefficients. This constraint would reduce the computational burden of inference while retaining the capability to handle complex output dependence. We next provide Markov chain Monte Carlo inference procedures for OSLMM and demonstrate superior performance on latent dynamics recovery in synthetic experiments of input-dependent Lorenz dynamics. We also show the superior computational benefits and prediction performance of OSLMM on several real-world applications. More importantly, we demonstrate the utility of OSLMM as a data analysis tool for neuroscience on two neural datasets of multi-neuron recordings. Both experiments demonstrate that OSLMM obtains superior latent representations which are better able to predict external variables and more accurately reflect the data generation process than GPFA. Together, these results demonstrate that OSLMM would be useful for the analysis of diverse, large-scale time-series datasets.

\section*{Author summary}
The utility of Gaussian processes has been demonstrated on a number of classic machine learning tasks, but providing scientific understanding of the result is not clear but desired. On the other hand, for high-dimensional systems, computation suffers from the ``curse of dimensionality", rendering the training unfeasible with a naive computing approach. In this study, our proposed OSLMM handles both two challenges by incorporating an interpretable orthogonal structure and proposing efficient inference procedures. We demonstrate its better recovery performance of latent trajectories in synthetic experiments and its computational benefits on various real datasets and the superior interpretable latent representations on two multi-neuron recording datasets. In the two neural datasets, we find that the latent representations extracted from our proposed OSLMM can be more accurately reflect the data generation process than those extracted from GPFA. It shows that our proposed OSLMM would be a very useful data analysis tool in large-scale neural time-series data and can provide important insight into neural responses.

%% file: INTRO.tex
\section*{Introduction}

Understanding the brain function is a central goal of neuroscience. Analysing the neuron response is one of the direct approach to understand the spatially distributed, dynamic patterns of brain activities. The main analysis approaches can be summarized into two classes. The first class of approaches is complex network analysis, analyzing the networks of brain regions connected by anatomical tracts or by functional association and describing important properties of complex system by quantifying topologies of network representations \cite{rubinov2010complex,yang2021disentangled}. And the other class is latent variable models, projecting high-dimensional neuron response to low-dimension space and exploiting the interpretation on the latent representations \cite{arieli1996dynamics, byron2009gaussian}.

In terms of latent variable models, recent works become increasingly appreciated that computations in the brain are carried out by the dynamics of neural populations. Thus, methods that extract the latent structure of those dynamics is critical to understanding brain function. The dynamics of population neural activity can result from both internal processing and external stimulus drive. In response to identical stimuli, the population dynamics can be different, as illustrated in a diversity of experiments. Therefore, instead of averaging noisy neuron records across multiple experimental trials, trial-by-trial based works on neuron response get increasing attention \cite{byron2009gaussian, wu2017gaussian, pandarinath2018inferring, she2020neural}. 

Trial-by-trials based works assume that the relevant population dynamics are often confined to a lower dimensional subspace and different approaches are proposed to model two mapping functions, the mapping function between time and latent trajectories and the other mapping function from latent trajectories to observations. One of the most popular approaches is Gaussian process factor analysis (GPFA) proposed by \cite{byron2009gaussian}. It models the neuron responses as a linear combination of independent Gaussian processes (GP), where the linear mapping models the correlation of neurons while GPs provide a flexible way to model latent trajectories. Moreover the GPs also impose the smoothness into latent trajectories which is important for interpretation. In general, the GPFA belongs to the linear model of coregionalization (LMC) \cite{bourgault1991multivariable,goulard1992linear} and several adaptions of LMC are capable of handling nonstationary covariances \cite{gelfand2004nonstationary,meng2021nonstationary}, datasets with large numbers of samples, and high-dimensional datasets \cite{zhang2021spatial,meng2021collaborative}. 

More recently, like GPFA, \cite{gao2015high,pandarinath2018inferring} use the linear coupling between latent dynamics and neural responses but model the dynamics using linear dynamic systems and recurrent networks, respectively. On the other hand, several studied have introduced nonlinear coupling between latent dynamics and neural responses, such as Gaussian processes \cite{wu2017gaussian,she2020neural}, neural nets\cite{gao2016linear}. Although those nonlinear mappings are flexible, they would impede geometric interpretation of the latent dynamics. We propose a general regression framework, the Orthogonal Stochastic Linear Mixing Model (OSLMM), in which an adaptive linear function (a conditional linear function) is employed as a nonlinear mapping. Moreover, instead of imposing orthogonality on coefficients \textit{post-hoc}, as in the GPFA, we directly impose orthogonality on stochastic coefficients. Both the conditional linear and orthogonal properties contributes to better visualization of latent dynamics.

Compared with other methods, some advantages and differences of the OSLMM are addressed. First, similar to most LMCs, the GPFA assumes fixed correlations between neurons, which is counter to the reality of neural dynamics. In neural data sets, it has been observed that there are time-dependent changes in correlation structure that are temporally aligned to the stimulus, supporting this assumption \cite{churchland2010stimulus,bouchard2016high}. We address this issue by employing an adaptive linear function of latent functions to allow the correlation structure changes over time. Recent work \cite{meng2021nonstationary} has also illustrated that the adaptive linear projection structure can deal with input-dependent correlation, scale and smoothness of outputs. Second, similar to the Orthogonal Instantaneous Linear Mixing Model (OILMM) in \cite{bruinsma2020scalable}, both models assume the coefficient matrices in the coupling between latent dynamics and neural responses are orthogonal. But the OSLMM assumes that the coefficient matrices vary across inputs while the OILMM does not. Lastly, we note that the OSLMM does not belong to multivariate Gaussian process model since the likelihood is non-Gaussian. 

In the context of neuroscience, it has been observed that there are time-dependent changes in correlation structure that are temporally aligned to the stimulus (e.g. \cite{churchland2010stimulus,dichter2016dynamic}). Thus it is necessary to release the time invariant correlation assumption in GPFA which is the state-of-the-art method in neural response modeling. Therefore, it illustrate that our proposed OSLMM is well-motivated for modelling  neural datasets.

In summary, we develop a new regression framework, the Orthogonal Stochastic Linear Mixing Model, for high-dimensional time series. We first develop Stochastic Linear Mixing Model (SLMM) where we employ Gaussian processes for dynamics and an adaptive linear function to model the coupling relation between latent dynamics and neural responses. Then we develop the OSLMM by putting an orthogonal structure on the adaptive coefficient matrices in the SLMM. We illustrate that the adaptive linear function and the orthogonal structure would contribute to high computational efficiency, superior predictive performance and insightful nonlinear interpretable dynamics from high-dimensional time series. Compared with the GPFA model, we show the better latent dynamics recovery performance on the synthetic experiments of input-dependent Lorenz Dynamics in different scenarios. Next, we derive the theoretical computational benefits and report empirical computational benefits in real datasets. We demonstrate the superior predictive performance on real datasets. Moreover, we find more insightful latent representations in applications to neurophysiology recordings from auditory cortex and motor cortex, respectively. Specifically, in the recording data of auditory cortex of rats, we show that the OSLMM subspaces exhibit monotonic ordering of stimulus amplitude and frequency, which is the expected organization given known auditory cortex response properties and in the recording data of motor cortex of monkeys, we show that the OSLMM subspaces exhibit monotonic ordering of reaching angles and velocities, which matches the expected organization given known motor cortex response properties. Such structure was not present in subspaces extracted by the GPFA. Thus, the OLSMM extracts latent structure from time-series data sets that provide greater insight into the neurobiological processes that generated the observed data.

%% file: METHOD.tex
\section*{Materials and Methods}

We propose the stochastic linear mixing model and orthogonal stochastic linear mixing model as well as their inference approaches. Then we describe the synthetic data generating process for input-dependent Lorenz dynamics for model validation and provide the details of analysis and evaluation metrics on the single-trial neural data.

\subsection*{Stochastic linear mixing model} \label{sec2:SLMM}

We first introduce a general class of Gaussian process based multivariate models called the stochastic linear mixing model (SLMM). Throughout this text we suppose $\bm y(\R{t}) \in \mathbb{R}^P$ be a vector-valued output function evaluated at the time input $\R{t}$,  where $P$ is the dimensionality of output. Given a dataset $\mathcal{D}$ of time inputs $\B{T} = [\R{t}_1, \ldots, \R{t}_T]$ and corresponding outputs $\B{Y} = [\B{y}_1, \ldots, \B{y}_T]$, we aim to predict $\bm y(\R{t}^*)| \R{t}^*, \mathcal{D}$ at a test input $\R{t}^*$, while accounting for input dependent signals across the elements of $\bm y(\R{t})$.

\textbf{Mod. 1} (Stochastic linear mixing model) Let $\bm f(\cdot) = \{f_1(\cdot), \ldots, f_Q(\cdot)\}$ be a vector-valued signal function composed of $Q$ independent latent functions. Each latent function is sampled from a GP prior with a squared exponential covariance function such that $f_q \sim \mathcal{GP}(0, k_{f})$ with $k_{f}(\R{t}, \R{t}) = 1$. $W(\R{t})$ is a $P\times Q$ input dependent coefficient matrix and $\Sigma$ is a $P\times P$ covariance matrix of observational noise. SLMM models the output function as a linear combination of latent functions corrupted with observation noises. Specifically, it is given by the following generative model:
\begin{align}
    f_q \stackrel{ind}{\sim} \mathcal{GP}(0, k_{f_q}) \,, \qquad \text{latent processes} \nonumber \\
    \bm g(\R{t})| W(\R{t}), \bm f(\R{t}) = W(\R{t}) \bm f(\R{t})\,, \qquad \text{mixing mechanism} \nonumber \\
    \bm y(\R{t}) | \bm g(\R{t}) \sim \mathcal{N}(\bm g(\R{t}), \Sigma) \,. \qquad \text{noise model} \nonumber 
\end{align}

We call $\bm f$ the latent processes and $W$ mixing coefficients. The SLMM is the generalization of the instantaneous linear mixing model (ILMM) \cite{bruinsma2020scalable}. Instead of employing a deterministic mixing coefficients $\B{W}$, the SLMM explicitly assumes that it depends on input $\R{t}$. This mixing mechanism with independent latent processes is called spatially varying linear model of corregionalization (SVLMC) \cite{gelfand2004nonstationary} in spatial statistics literature. Recently, \cite{meng2021nonstationary} propose a general regression framework based on this mixing mechanism and get a successful implementation of the analysis in electronic health records. On the other hand, replacing latent processes $\bm f(\R{t})$ with noisy latent processes $\bm f(\R{t}) + \sigma_f \bm \epsilon$, assuming homogeneous noise such that $\Sigma = \sigma_y^2 I_P$ and modeling each element of $W(\R{t})$ via a Gaussian process lead the SLMM to be the exact Gaussian process regression network (GPRN) in \cite{wilson2011gaussian}. 

Following \cite{wilson2011gaussian}, we assume all the latent functions share the same covariance function $k_f$, and also assume that the function of each mixing coefficient $w_{ij}(\R{t})$ is independently sampled from a GP with the same covariance function $k_w$. We denote the values of $f_q$ at inputs $\B{T} = [\R{t}_1, \ldots, \R{t}_T]'$ by $\B{f}_{q,\cdot} = [f_q(\R{t}_1), \ldots, f_q(\R{t}_T)]'$, the values of $w_{ij}$ at inputs $\B{T}$ by $\B{ w}_{ij} = [w_{ij}(\R{t}_1), \ldots, w_{ij}(\R{t}_T)]'$. The joint probability of observed outputs $\B{Y} = [\B{y}_1, \ldots, \B{y}_T]$ and latent variables $\{\B{w}_{ij}\}$ and $\{\B{f}_{q,\cdot}\}$ is
\begin{align}
    p(\B{Y}, \{\B{w}_{ij}\}, \{\B{f}_q\}| \B{X}, \theta_f, \theta_w, \Sigma) = \prod_{t = 1}^T\mathcal{N}(\B{y}_t| \B{W}_t \B{f}_t, \Sigma) \prod_{i = 1}^P\prod_{j = 1}^Q\mathcal{N}(\B{w}_{ij}| \B{0}, \B{K}_w)\prod_{q = 1}^Q\mathcal{N}(\B{f}_{q,\cdot}| \B{0}, \B{K}_{f}) \label{eq:joint_SLMM}
\end{align}
where $\B{W}_t$ is a $P \times Q$ coefficient matrix at time $t_t$ in which $[\B{W}_t]_{ij} = w_{ij}(\R{t}_t)$, $\B{f}_t = \bm f(\R{t}_t)$. $\B{K}_w$ and $\B{K}_{f}$ are the covariance matrices estimated at inputs $\B{T}$, and model parameters are $\Theta = (\theta_f, \theta_w, \Sigma)$.

Learning in SLMM is equivalent to inference of the posterior distribution of latent variables and model parameters. Latent variables consist of mixing coefficients ($\{\B{w}_{ij}\}$) and latent variables ($\{\B{f}_{q,\cdot}\}$), and model parameters include the covariance matrix of observation noise $\Sigma$ and hyper-parameters in GPs. The most computationally expensive component of the learning procedure comes from inference of latent variables. We note that the conditional posterior of mixing coefficients $p(\B{W}|\B{f}, \B{Y}, \B{T},\ \theta_f, \theta_w, \Sigma)$ and the conditional posterior of latent functions $p(\B{f}| \B{W}, \B{Y}, \B{T},\ \theta_f, \theta_w, \Sigma)$ have close-form expressions. They are multivariate Gaussian distributions with dimension $PQT$ and $QT$. The complexity of $learning$ them are $\mathcal{O}(P^3Q^3T^3)$ and $\mathcal{O}(Q^3T^3)$ respectively. Hence, the Gibbs sampling for $\B{W}$ and $\bm f$ would be difficult for large datasets. \cite{wilson2011gaussian} propose a Markov-chain Monte-Carlo (MCMC) approach to jointly sample them via elliptical slice sampling (ESS), an acceptance-rejection sampling method \cite{murray2010elliptical}. The time complexity of ESS depends on that of computing the joint distribution of (\ref{eq:joint_SLMM}) which takes $\mathcal{O}(PQT^3)$ (shown in the supplementary of \cite{wilson2011gaussian}). Although the ESS relieves the computational burden, ESS still does not work for large datasets in practice, because of the poor mixing.  

Our inference conditionally samples latent variables $\B{W}$ and $\B{f}$ given model parameters $\Theta$ via ESS and conditionally samples $\Theta$ given $\B{W}$ and $\B{f}$. The details of sampling model parameters are described in \nameref{S1_Appendix}. Similar to the inference in \cite{wilson2011gaussian}, our inference is not efficient, because that ESS suffers from the low efficiency and slow time to convergence. Therefore, we next propose a new regression framework, the orthogonal stochastic linear mixing model that introduces an orthogonal constraint amongst the mixing coefficients and significantly improves the inference efficiency theoretically and empirically. 

\subsection*{Orthogonal stochastic linear mixing model}\label{sec3:OSLMM}
In SLMM, the most burdensome computation comes from the inference of mixing coefficients $\B{W}$, which includes $PQN$ model parameters. To improve the inference efficiency, we simplify the model by introducing an orthogonal constraint amongst the mixing coefficients. We call this new model the orthogonal stochastic linear mixing model (OSLMM).

\begin{figure}[ht!]
    \centering
    \includegraphics[width=\textwidth]{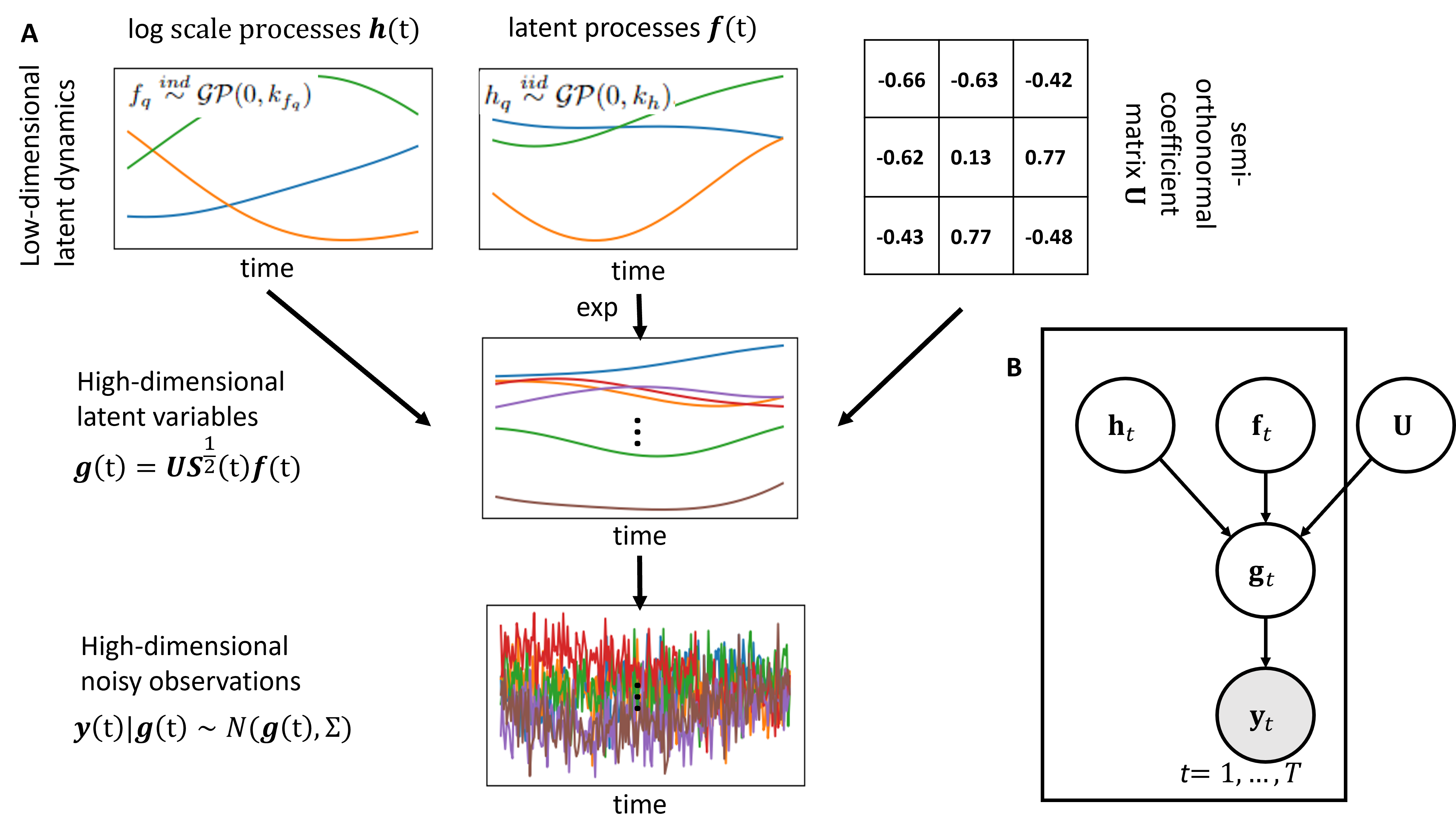}
    \caption{Schematic diagram of the Orthogonal Stochastic Linear Mixing Model (OSLMM). Panel A illustrates data generated by the model with a three-dimensional latent processes. Panel B refers to the illustration of the graphical model of OSLMM.}
    \label{fig:graphical_model}
\end{figure}

Instead of explicitly modeling the mixing coefficients $W(\R{t})$ via GPs, OSLMM takes the eigen-decomposition on the variance-covariance matrix of the latent signal $\bm g(\R{t})$ given $W(\R{t})$, implying that $\R{var}(\bm g(\R{t})) = W(\R{t})W(\R{t})' = U(\R{t})S(\R{t})U(\R{t})'$, where the columns of $U(\R{t}) \in \mathbb{R}^{P \times Q}$ are orthonormal and $S(\R{t}) \in \mathbb{R}^{Q \times Q}$ is a positive diagonal matrix. Then $W(\R{t})$ can be decomposed as $W(\R{t}) = U(\R{t})S^{\frac{1}{2}}(\R{t})$. We simplify the structure of mixing coefficients by assuming $U(\R{t})$ is independent from input $\R{t}$: $W(\R{t}) = \B{U}S^{\frac{1}{2}}(\R{t})$. Then the latent signals $\{\bm g(\R{t})\}$ would stay in the subspace spanned by the orthonormal basis of $\B{U}$. This assumption is accordance with the observation that high dimensional data usually lie on a low-dimensional manifold in many real-world problems \cite{bouchard2013functional}. Specifically, the model is:

\textbf{Mod. 2} (Orthogonal stochastic linear mixing model) The OSLMM is an SLMM (\textbf{Mod. 1}) where the latent signal $\bm g(\R{t})$ is expressed as $\bm g(\R{t}) = W(\R{t})\bm f(\R{t}) = \B{U}S^{\frac{1}{2}}(\R{t})\bm f(\R{t})$ where $\B{U}$ is a $P \times Q$ matrix with orthonormal columns, $S(\R{t})$ is a $Q \times Q$ positive diagonal matrix indexed by input $\R{t}$, and $\Sigma = \sigma^2_y\B{I}$. 

In order to model the positive diagonal matrices $\{S^{\frac{1}{2}}(\R{t})\}$,  OSLMM assumes that each element on the diagonal of $S^{\frac{1}{2}}(\R{t})$ in the logarithmic scale, $h_q(\R{t}) = \log([S^{\frac{1}{2}}(\R{t})]_{qq}$, has a GP prior with a squared exponential covariance function such that $h_q \stackrel{iid}{\sim} \mathcal{GP}(0, k_h)$. We display the schematic diagram of OSLMM in Figure~\ref{fig:graphical_model} as well as the corresponding graphical model. In the schematic diagram, the latent dimension size is $Q=3$. 

In the comparison with SLMM, the number of latent variables of OSLMM is reduced from $PQT + QT$ to $PQ + 2QT$. In practice, this reduction in parameters renders inference possible for large datasets. In addition, we develop an efficient inference framework via sufficient statistics as follows.

Similar to \cite{bruinsma2020scalable}, we first propose projection matrices $\{\B{T}_t\}$ such as $\B{T}_t = \B{S}_t^{-\frac{1}{2}}\B{U}'$, where $\B{S}_t= S(\R{t}_t)$. Conditional on $\B{U}, \B{S}_t$, $\B{T}_t\B{y}_t$ is a maximum likelihood estimate for $\B{f}_t$. In addition, $\B{T}_t\B{y}_t$ is a minimally sufficient statistic for $\B{f}_t$. The detailed proofs are provided in \nameref{S2_Appendix}. Those summary statistics lead to the fact that for any prior $p(\B{f}_t)$ over $\B{f}_t$, we have
\begin{align}
    p(\B{f}_t| \B{y}_t) = p(\B{f}_t| \B{T}_n\B{y}_t)\,,\qquad
    \B{T}_t\B{y}_t| \B{f}_t \stackrel{ind}{\sim} \mathcal{N}(\B{T}_t\B{y}_t| \B{f}_t, \Sigma_{T_t}) \label{eq:suffiency} 
\end{align} 
where $\Sigma_{T_t} = \B{S}^{-\frac{1}{2}}_t\B{U}'\Sigma\B{U}\B{S}^{-\frac{1}{2}}_t$. When $\Sigma$ has the form $\Sigma = \B{U}\B{D}_1\B{U}' + \sigma^2_y\B{I}$, the variance-covariance matrix is a diagonal such that $\Sigma_{T_t} = \B{S}^{-\frac{1}{2}}_t \B{D}_1\B{S}^{-\frac{1}{2}}_t + \sigma^2_y\B{S}_t^{-1}$. It would contribute to a linear learning complexity with respect to latent functions $\B{f}$ in \eqref{eq:cond_f}. In the following of this work, we assume a homogeneous noise $\Sigma = \sigma^2_y\B{I}$, and thus $\{\Sigma_{T_t}\}$ are diagonal.

Further, we refer to $\bm c(\R{t}) = S^{\frac{1}{2}}(\R{t})\bm f(\R{t})$ as the orthonormalized latent functions. Each dimension of $\bm c(\R{t})$ represents a scaled $\bm f(\R{t})$ at each input $\R{t}$. Similar to the orthonormalized neural state in GPFA \cite{byron2009gaussian}, the orthonormalized latent functions can explain the amount of data covariance. Also, similar to the spirit in PCA \cite{yu2009gaussian,vyas2020computation}, this orthonormality constraint penalizes redundant latent representation \cite{salzmann2010factorized} and then contributes to a better low-dimensional visualizations of the latent structure. We also note that the GPFA impose the orthogonality after inference, which may lead to mixing of data effects in latent factors that can not demixed by post-hoc orthogonalization in an unsupervised manner. In OSLMM, this orthogonality constraint is imposed directly during inference, allowing it to capture orthogonal structure in the data directly. That contributes to better interpretation on latent trajectories.

We propose a Markov chain Monte Carlo (MCMC) algorithm for OSLMM via Gibbs sampling, which updates latent functions and model parameters iteratively from their conditional posterior distributions. First, because of (\ref{eq:suffiency}), the conditional posterior of latent variables $\B{f}$ is 
\begin{align}
    p(\B{f} | \B{H}, \B{S}, \B{Y}, \B{X}, \theta_f, \theta_w, \Sigma) & \propto \prod_{t = 1}^T\mathcal{N}(\B{T}_t\B{y}_t| \B{f}_t, \Sigma_{T_t}) \prod_{q = 1}^Q\mathcal{N}(\B{f}_{q,\cdot}| \B{0}, \B{K}_{f}) \nonumber \\
    & = \prod_{q=1}^Q \mathcal{N}(\B{f}_{q, \cdot}|(\B{K}_{f}^{-1} + \tilde{\Sigma}_q^{-1})^{-1}(\tilde{\Sigma}_q^{-1}\tilde{\B{y}}_q) , (\B{K}_{f}^{-1} + \tilde{\Sigma}_q^{-1})^{-1})\,, \label{eq:cond_f}
\end{align}
where $\tilde{\Sigma}_q = \mathrm{diag}([\Sigma_{T_1}]_{qq}, \ldots, [\Sigma_{T_T}]_{qq})$ and $\tilde{\B{y}}_q = ([\B{T}_1\B{y}_1]_q, \ldots, [\B{T}_T\B{y}_T]_q)'$.

Because this conditional posterior can be factorized into the product of each latent dimension $q$, and each conditional posterior is a multivariate Gaussian distribution, the learning complexity is $\mathcal{O}(T^3Q)$, linear to the latent dimension size $Q$. The conditional posterior of $\B{h} = (\B{h}_1, \ldots \B{h}_Q)$, where $\B{h}_q = (h_q(\R{t}_1), \ldots, h_q(\R{t}_T))'$, is 
\begin{align}
    p(\B{h} | \B{H}, \B{f}, \B{Y}, \B{X}, \theta_f, \theta_w, \Sigma) & \propto \prod_{t = 1}^T\mathcal{N}(\B{y}_t| \B{U}\B{S}^{\frac{1}{2}}_t\B{f}_t, \Sigma) \prod_{q = 1}^Q\mathcal{N}(\B{h}_q| \B{0}, \B{K}_h) \nonumber \\
    & \propto \prod_{t=1}^T \exp(-\frac{1}{2} (\B{y}_t - \B{U}\B{S}^{\frac{1}{2}}_t\B{f}_t)' \Sigma^{-1} (\B{y}_t - \B{U}\B{S}^{\frac{1}{2}}_t\B{f}_t) ) \prod_{q = 1}^Q\mathcal{N}(\B{h}_q| \B{0}, \B{K}_h)\,.
\end{align}

As $\Sigma$ is diagonal, this likelihood can be factorized for each time index $t$ and each output dimension $p$. So the computational complexity of this posterior is $\mathcal{O}(\max(PT, T^3))$. Since the closed-form expression of each posterior is intractable, we sample them via the elliptical slice sampling \cite{murray2010elliptical}. 

To sample $\B{U}$, because $\B{U}$ is on the Stiefel manifold where the columns of it are orthonormal, we parametrize $\B{U}$ with the polar decomposition such that $\B{U} \stackrel{d}{=} \B{U}_{\B{V}} = \B{V}(\B{V}^T\B{V})^{-\frac{1}{2}}$ \cite{jauch2020monte}, where $\B{V}\in\mathbb{R}^{P\times Q}$ is a random matrix. We assume $p_\B{U}(\B{U})$ is uniform and thus $\B{V}$ follows a matrix angular central Gaussian distribution, $\R{MACG}(\B{I}_P)$, corresponding to $\B{V} \sim \mathcal{N}_{P,Q}(\B{0}, \B{I}_P, \B{I}_Q)$ \cite{chikuse2012statistics}. Hence, the conditional posterior of $\B{V}$ is 
\begin{align}
    p(\B{V}|\B{f}, \B{S},  \B{Y}, \B{T}, \theta_f, \theta_w, \Sigma) & \propto \prod_{t = 1}^T\mathcal{N}(\B{y}_t| \B{U}\B{S}^{\frac{1}{2}}_t\B{f}_t, \Sigma) \mathcal{N}_{P,Q}(\B{V}|\B{0}, \B{I}_P, \B{I}_Q)  \nonumber \\
    & \propto \prod_{t=1}^T \exp(-\frac{1}{2} (\B{y}_t - \B{U}\B{S}^{\frac{1}{2}}_t\B{f}_t)' \Sigma^{-1} (\B{y}_t - \B{U}\B{S}^{\frac{1}{2}}_t\B{f}_t) ) \mathcal{N}_{P,Q}(\B{V}|\B{0}, \B{I}_P, \B{I}_Q)\,.
\end{align}
We sample $\B{V}$ via elliptical slice sampling and the computational complexity of this posterior is $\mathcal{O}(\max(PT, PQ))$. 

\begin{figure}[ht!]
    \centering
    \includegraphics[width=2in, height=2in]{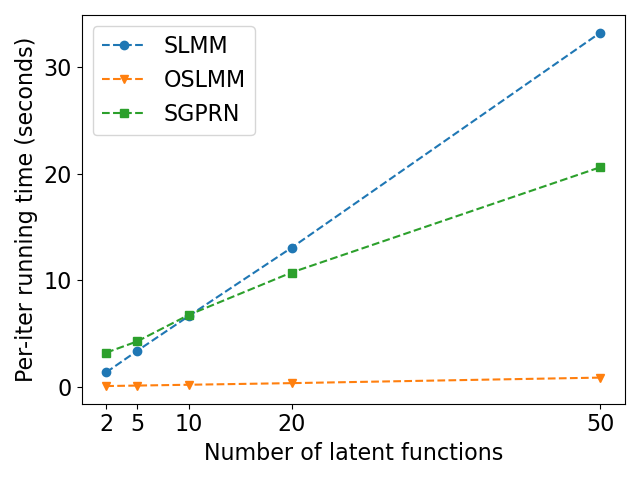}
    \caption{Training speed of SLMM, OSLMM and SGPRN inference algorithms on \textbf{Neural} data. We show the running time per iteration in the setting with different number of latent functions.}
    \label{fig:running time}
\end{figure}

Finally, to update model parameters $\Theta$, we employ the Metropolis Hasting method and the details are discussed in \nameref{S3_Appendix}. This inference takes $\mathcal{O}(\max(QT^3, PT, PQ))$ time, which is linear in the number of latent dimensions $Q$ and output variable dimensionality $P$. Empirically, we compare the training speed of OSLMM to that of SLMM and sparse Gaussian process regression network (SGPRN) \cite{li2020scalable} in one neural dataset with output dimension $128$. This experiment takes 100 time stamps for training and we report the running time for each iteration in Figure~\ref{fig:running time}. These results clearly demonstrate that inference of OSLMM significantly faster than SLMM and SGPRN. The details of data and methods are available in \nameref{S4_Appendix}. In \nameref{S4_Appendix}, we display the same training speed comparison on two other real high-dimensional machine learning datasets. Moreover, we report the predictive performance on five real data in \nameref{S4_Appendix}, which shows that OSLMM has better predictive performance on most of datasets.

\subsection*{Data generating process from input-dependent Lorenz dynamics}

We recover the well-known input-dependent Lorenz dynamics in a nonlinear system, the input-dependency refers to that the Lorenz dynamics is scaled by time-varying scale factors. Specifically, the Lorenz system describes as a two dimensional flow of fluids with $f_q$, $q=1,2,3$ as latent processes.
\begin{align}
    \frac{df_1}{d t} = \sigma(f_2 - f_1), \frac{d f_2}{d t}=f_1(\rho-f_3) - f_2, \frac{df_3}{dt} = f_1 y- \beta f_3\,. \label{eq:lorenz}
\end{align}
Lorenz sets the values $\sigma=10, \beta=8/3$ and $\rho=28$ to exhibit a chaotic behavior as the same utilized in recent works \cite{she2020neural,zhao2017variational,linderman2017bayesian}. 

We simulated a three-dimensional latent dynamics using Lorenz system in \eqref{eq:lorenz} and normalized each dimensions with unit variance and zero mean. Then we apply three different mapping functions of log scales such that $h_q \stackrel{iid}{\sim} \mathcal{GP}(0, k_h)$ with different squared exponential covariance functions, $k_h^{\text{short}}(\Delta t) = \exp(-\frac{\Delta t^2}{2})$, $k_h^{\text{median}}(\Delta t) = \exp(-\frac{\Delta t^2}{2\exp(1)^2})$, and $k_h^{\text{long}}(\Delta t) = \exp(-\frac{\Delta t^2}{2\exp(2)^2})$. Finally, we choose a random semi-orthogonal matrix $\B{U}$ and generate data via $\bm y_t = \B{U}^T \exp({\B{h}}_t) \B{f}_t + \B{\eta}_t$, where $\B{h}_t = \bm{h}(t_t)$ and the noise $\B{\eta}_t$ are drawn from $\mathcal{N}(\B{0}, 0.1^2\B{I})$.

\subsection*{Analysis of single-trial neural data} \label{sec:single_trial_analysis}

Inferring latent trajectories, particularly from single trial neural population recordings, may help us understand the dynamics that produce brain computations \cite{vyas2020computation}. A large class of methods assumes an autoregresive linear dynamics model in the latent process due to the computational feasibility \cite{paninski2010new,kao2015single}. However, the assumption of linear dynamics may be overly simplistic since interesting neural computations are naturally nonlinear in the brain in general. Therefore Gaussian process factor analysis method (GPFA) is proposed \cite{byron2009gaussian,lakshmanan2015extracting}. Similar to GPFA, OSLMM imposes a general Gaussian process prior to infer latent dynamics. However, OSLMM differs from GPFA in three aspects. First, OSLMM assumes that the coefficient matrix $W(\R{t})$ is time dependent, which allows modelling time-varying correlation across neurons/channels. This is critical, as it is known that the correlation structure of neural data changes over time. Second, GPFA orthogonalisation of the columns of coefficient matrix $W$ is done as a post-processing step while OSLMM builds the orthogonalisation of $W(\R{t})$ into the model, arguably a more desirable modeling approach. Finally, GPFA provides only point estimates of values, while OSLMM provides samples from the posterior distribution. In the single-trial neural analysis, we conduct analysis on two datasets. 

\subsubsection*{Analysis on rat auditory cortex experiments}

One dataset comes from rat auditory cortex experiments. We collected micro-electrocorticography ($\upmu$ECoG) data in the rat auditory cortex experiments in the Bouchard Lab \cite{dougherty2019laminar}. We analyzed the z-scored high-gamma activity of $128$ simultaneously recorded $\upmu$ECoG channels over rat auditory cortex. High-gamma (70-170Hz) activity from $\upmu$ECoG is a commonly-used signal containing the majority of task relevant information for understanding the brain computations  \cite{livezey2019deep}. For each experimental trial, we analyzed neural activity for a duration of $150$ ms in which the auditory stimuli happened from $50$ ms to $100$ ms. The stimuli consisted of $240$ different sounds with $8$ distinct amplitudes from $1$ to $8$ [-70 to 0 dB attenuation] and $30$ distinct frequencies from $\SI{500}{\hertz}$ to $\SI{32000}{\hertz}$. \cite{dougherty2019laminar}. Each stimulus has $20$ trials in the experiment. The neural activity was downsampled to $\SI{400}{\hertz}$.  We calculated leave-one-channel-out prediction error (accuracy), and additionally explored the latent representation of the data. 

As for the electrocorticogrphy data (ECoG), we conducted both stimuli-wise and global analysis. Specifically, \SI{}{\micro\metre} ECoG neural recordings come from rat auditory cortex in response to multiple stimuli, and each stimulus is presented on multiple randomly interleaved trials. Each trial includes a multivariate time series (the z-scord high-gamma band amplitude across \SI{}{\micro\metre} ECoG electrodes). In stimuli-wise analysis, we assume that within signal stimuli the mixing coefficients $\B{W}$ are shared across all trials and different trials have their own individual latent processes. And the mixing coefficients $\B{W}$ are shared across all trials and stimuli in the global analysis. 

\subsubsection*{Analysis on monkey arm-reaching experiments}

The second dataset is obtained from the monkey arm-reaching experiments and comes from \cite{churchland2010stimulus,pei2021neural}. It consists of one full session with 2869 total trials (2295 trials for training and 574 trials for testing), 108 conditions and 182 neurons with simultaneously monitored hand kinematics. The arm-reaching task is a delayed center-out reaching task, including three task timelines: target presentation, go cue and movement onset. We aligned the neural data from 50 ms before the move onset time to 450 after that and resampled the data at the bin size 5ms. Therefore, each trial has a multivariate spike time series with 100 time stamps. We smoothed spikes with Gaussian filter with 50 ms standard deviation.

As for the monkey arm-reaching data, we conducted the global analysis where the mixing coefficients are shared across all trials and conditions.

\subsubsection*{Model evaluation using leave one channel prediction}

For model evaluation, the leave-one-channel-prediction is considered for model comparison. We use three-fold cross-validation of all trials and so we have three pairs of training trials and testing trials. For each pair of data, we infer the posterior samples (OSLMM) or point estimates (GPFA) of shared latent variables $\B{U}$ and $\B{h}$, and model parameters $\Theta$ from training trials. Next, for each test trail, we leave one channel out of the test trial as a target neuron and compute the posterior predictive mean of the signal of the target channel using the remaining channels with the posterior samples (OSLMM) or estimates (GPFA) of shared latent variables and model parameters from the training trials. We repeat this procedure on each test trial and each channel of the chosen test trial. Finally, we choose the sum of square error as prediction error and coefficient of determination ($R^2$) as two prediction measures for model comparison.   
 
As single-trial neural data are regularly sampled in time, a convariance matrix generated from a stationary kernel has a Toeplitz structure. Specifically, for any Toeplitz matrix $\R{S} \in \mathbb{R}^{T \times T}$ with constant diagonals and $\R{S}_{i,j} = \R{S}_{i+1, j+1}$, this structure of a covariance matrix allows the corresponding GP inference in $\mathcal{O}(T\log T)$ and the GP prediction on variance in $\mathcal{O}(T^2)$ \cite{cunningham2008fast,  wilson2015thoughts}. Therefore, the learning complexity for our MCMC algorithm for single-trail data would be decreased to $\mathcal{O}(\max(QT\log T, PT, PQ))$.

We evaluated OSLMM on real benchmark datasets and analyzed single-trial neural data. Experiments are run on Ubuntu system with Intel(R) Core(TM) i7-7820X CPU @ 3.60GHz and 128G memory.

%% file: RESULT.tex
\section*{Results} \label{sec:experiments}

We provided three experiments in this section. We first show the superior recovery performance on input-dependent Lorenz dynamics synthetic experiments. And then we conducted analysis on two neuroscience experiments. The first analysis is on rat auditory cortex experiments \cite{dougherty2019laminar}. The data consists of micro-electrocorticography high gamma response to tone pips of varying frequency and attenuation. We showed that OSLMM achieves better prediction performance than GPFA and OSLMM extracts more interpretable representations than GPFA. The second analysis is on monkey arm-reaching experiments \cite{churchland2012neural,pei2021neural}. The data consist of simultaneously recordings from primary motor and dorsal cortices while a monkey make reaches with an instructed delay to visually presented targets while avoiding the boundaries of a virtual maze. We showed that OSLMM extracts more interpretable representations than GPFA and OSLMM's latent representation are more predictive to behavior than GPFA.

\subsection*{OSLMM provides superior recovery performance on input-dependent Lorenz dynamics synthetic experiments}

We recovered the Lorenz Dynamics from the synthetic data. Our simulation takes the latent dimension $Q=3$ and the number of neurons $P = 50$. We tested the ability of GPFA and OSLMM method to infer the latent input-dependent dynamics of the Lorenz dynamics from observation generated from the data generating processes (DGP). The details of DGP are provided in Section Methods. We compared the performance of latent trajectories reconstruction using the root mean squared error (RMSE) and displayed the difference of RMSEs between GPFA and OSLMM, $\Delta\text{RMSE} = \text{RMSE}_{\text{GPFA}} - \text{RMSE}_{\text{OSLMM}}$. The larger $\Delta\text{RMSE}$ is, the better OSLMM performs than GPFA. We considered three different scenarios. 

\begin{figure}[ht!]
    \centering
    \includegraphics[width=\linewidth]{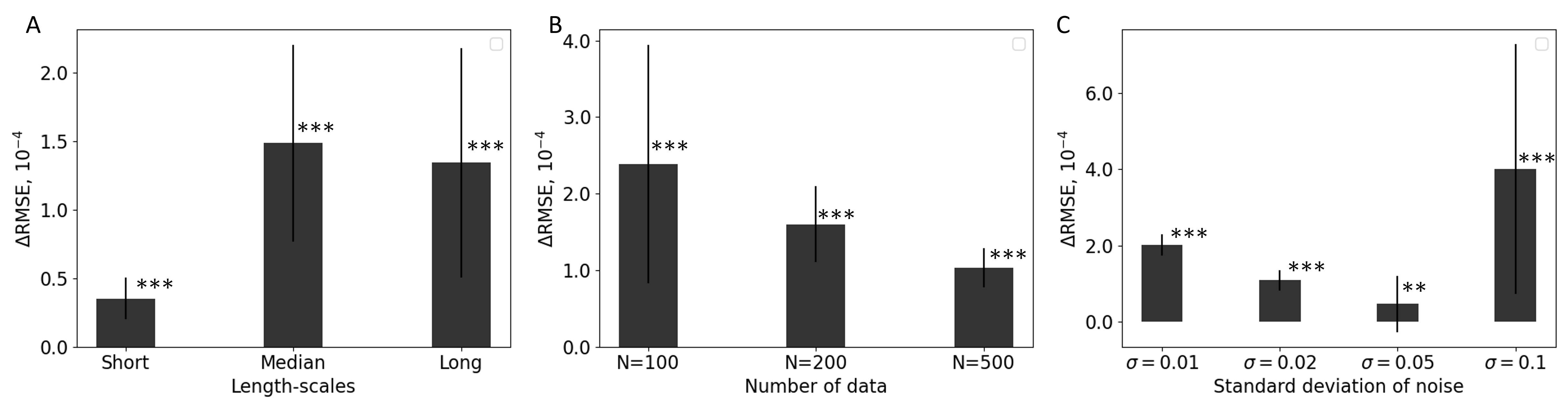}
    \caption{OSLMM provides superior recovery performance on input-dependent Lorenz dynamics synthetic experiments: The difference of root mean square error (RMSE, $10^{-4}$) of latent trajectories reconstructed from GPFA and OSLMM
    in three different scenarios. (A) Different scales: Three data generation processes (DGP) with different scale function generated by log Gaussian process in terms of short, median or long length-scales an we set data size $N = 200$. (B) Different data size: Three median length-scale DGPs in terms of various number of time stamps, N=100, 200, 500. (C) Different standard deviation of noise: Four MDGPs in terms of different levels of noise on the subspace specified by the standard deviation of noise $\sigma$. The results in (A) and (B) are conducted over 10 trials and the results in (C) are conducted over 20 trials. All difference of RMSEs are summarized by mean and standard deviation. Moreover, we conducted a pair T-test on the difference of RMSEs and annotated the significance level for each case. The significance level is defined via the p value such that $\text{*}: 0.01<p\leq 0.05$, $\text{**}: 0.001<p\leq 0.01$ and $\text{***}: p \leq 0.001$.}
    \label{fig:LORENZ_RES}
\end{figure}

In the first scenarios, we 
took three DGPs with short, median and long length-scales in corresponding Gaussian processes. Each setting consists of 10 trials and we plotted the $\Delta\text{RMSE}$ in three cases summarized by the mean and standard deviation over the 10 trials in Figure~\ref{fig:LORENZ_RES}A. In addition, we conducted a paired T-test on the difference of RMSEs between GPFA and OSLMM and the p-values for three settings are $7.03\times 10^{-5}, 1.58\times 10^{-4}$ and $9.61\times 10^{-4}$. All results show that OSLMM achieves better recover performance on the latent trajectories than GPFA in all settings. The result of p-values indicates that the less smooth scales (shorter length-scale) lead to more significant difference between the performance of GPFA and OSLMM .

We evaluated the relation between the recovery performance and the number of data in  Figure~\ref{fig:LORENZ_RES}B. Specifically, we conducted the comparison between GPFA and OSLMM under different number of data for training (N=100, 200, 500) using the same DGP with median length-scales Guasisan processes. We also plotted the $\Delta\text{RMSE}$ in three cases summarized by the mean and standard deviation over the 10 trials and also conducted a paired T-test with p-values $1.29\times 10^{-3}, 4.91 \times 10^{-6}$ and $6.73 \times 10^{-7}$ for $N = 100, 200$ and $500$ settings. Both $\Delta\text{RMSE}$ and p-values show that OSLMM provides a superior recovery performance than GPFA. And the p-values suggests that the larger number of data leads to more significant difference between the performance of GPFA and OSLMM.

On the other hand, in neural experiments, each trial of neural responses may have different latent subspaces. Therefore, we used a different DGP for multiple trials called multiple data generating process (MDGP). We first generated one random semi-orthogonal matrix $U$, generated $20$ corrupted semi-orthogonal matrix $U_i$ by simulating $V_i = U + \sigma E_i, E_i \sim \mathcal{MN}(0 , I_P, I_Q)$ and next extracted the closest matrix in $\mathcal{V}(P, Q)$ to $V_i$ in the Frobenius norm, i.e. $U_i=\text{argmin}_{U\in \mathcal{V}(P, Q)} \|V_i - U\|_F$, where $\mathcal{V}(P, Q) = \{A \in \mathbb{R}^{P\times Q}| A^TA = I\}$ is the set of semi-orthogonal matrices. It suggests that $U$ is the median subspace of $\{U_i\}$. In this data generation process, we take $\sigma = 0.01, 0.02, 0.05, 0.1$, representing different levels of discrepancy in subspaces. We choose the number of data $N=200$ and the median length-scale GP for the mapping function of log scales. Finally, we compared GPFA and OSLMM on the latent trajectory reconstruction via the mean and standard deviation of $\Delta\text{RMSE}$ in Figure~\ref{fig:LORENZ_RES}C. We also conducted a paired T-test with p-values $7.72\times 10^{-18}, 3.26\times 10^{-13}, 1.33\times 10^{-2}, 3.82\times 10^{-5}$ for $\sigma = 0.01, 0.02, 0.05, 0.1$. Both results also that OSLMM consistently outperforms GPFA on the recovery performance for all levels of discrepancy in subspaces in the MDGPs.

\subsection*{Analysis on rat auditory cortex experiments}

We applied GPFA and OSLMM for both stimuli-wise and global analysis on rat auditory cortex experiments. We illustrate that OSLMM provides better predictive performance in term of lower leave-one-channel-out prediction error and more interpretable latent representations than GPFA in the sense of providing trajectories which can better reflect expected distributed auditory cortical population response properties.

\begin{figure}[ht!]
    \centering
    \includegraphics[width=\textwidth]{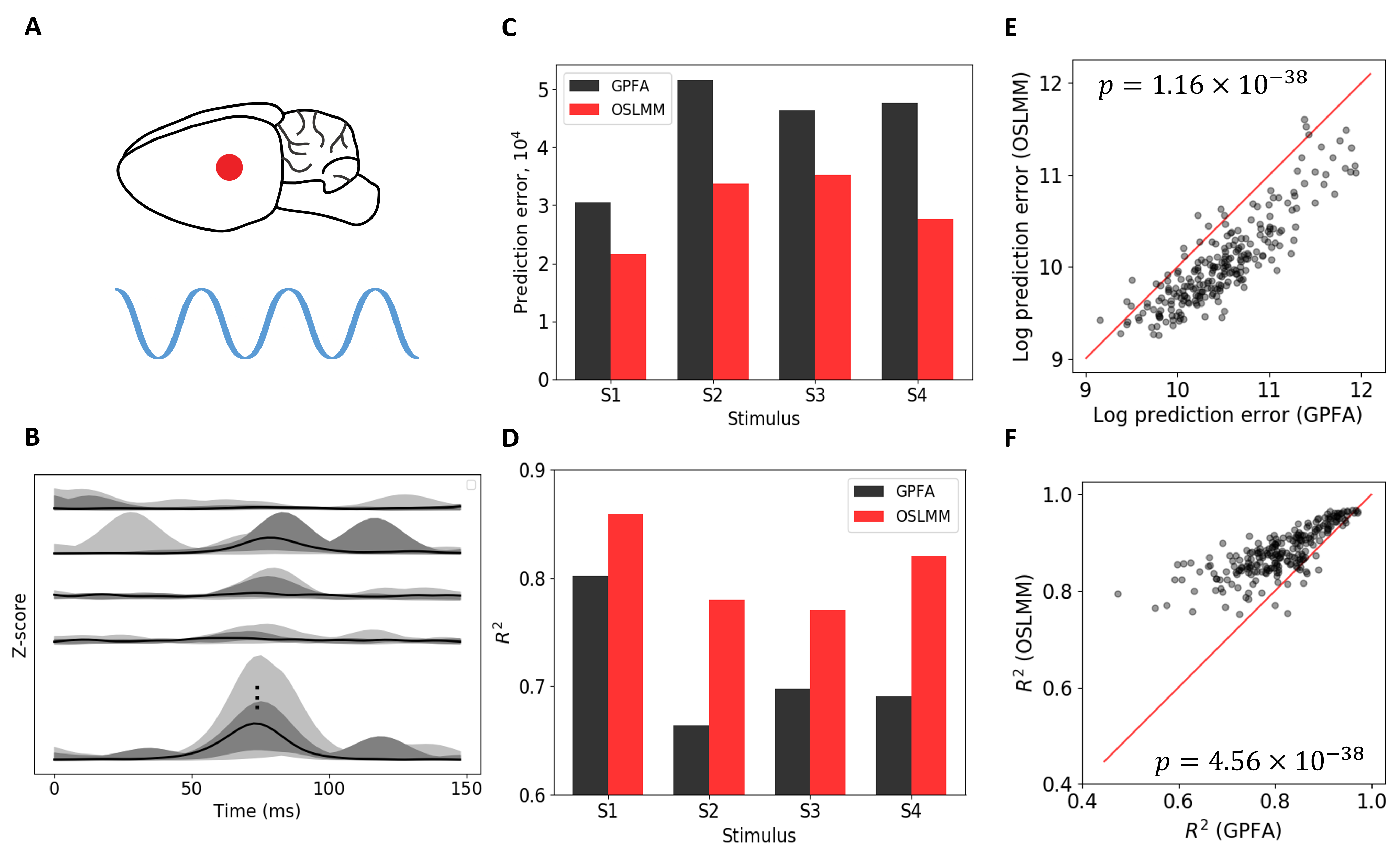}
    \caption{OSLMM achieves superior prediction performance for neural responses on holdout channels: Neural data and prediction performance on leave-one-channel out task. (A) Neural response from rat auditory cortex. (B) Functional boxplot of the Z-score curves for a stimuli where black lines refer to the median curve and light/dark grey shaded areas refer to functional data envelopes and $50\%$ central region. The stimulus takes frequency 32000 Hz and attenuation -10 dB and it starts from 50 ms and ends at 100 ms. (C-D) Prediction error (sum of square error) in (C) and coefficient of determination $(R^2)$ in (D) on the stimuli-wise analysis for the four stimuli. (E-F) Stimuli-wise log prediction error (smaller is better) (E) and coefficient of determination (larger is better) (F) on the the global analysis across all stimuli. We also provided the p values for the Wilcoxon sign-ranked test on the log prediction error and $R^2$ and they illustrates that OSLMM significantly outperforms GPFA in terms of prediction.}
    \label{fig:prediction_ECoG}
\end{figure}

\subsubsection*{OSLMM achieves superior prediction performance for neural responses on holdout channels} \label{sec:lono_analysis}

To quantify the predictive performance of GPFA and OSLMM on micro ECoG data, we used the prediction error and coefficient of determination $(R^2)$ (commonly used in neuroscience) in a leave-one-channel-out procedure as described in Section Methods. A smaller prediction error implies better prediction while a higher coefficient of determination implies better prediction. 
We conducted the stimuli-wise analysis by choosing four stimuli $\R{S}1$, $\R{S}2$, $\R{S}3$ and $\R{S}4$. $\R{S}1$ and $\R{S}2$ have the same attenuation $-10$ dB, and $\R{S}3$ and $\R{S}4$ have the same attenuation $-50$ dB. $\R{S}1$ and $\R{S}3$ have the same frequency $\SI{7627}{\hertz}$, and $\R{S}3$ and $\R{S}4$ have the same frequency $\SI{32000}{\hertz}$. 

The neural activities are summarized by the Z-score curves from a rat's auditory cortex as in Figure~\ref{fig:prediction_ECoG}A. We visually summarized the neural  activities via the functional boxplot \cite{sun2011functional,meng2021nonstationary} of the Z-score curves within each channel. We took the data from the stimulus $\R{S}1$ for example and visualized them in
Figure~\ref{fig:prediction_ECoG}B. The solid black line denotes the median curve, and the light/dark regions areas refer to the functional data envelopes and $50\%$ central region. The panel shows that stimulus would cause larger neural responses. 

We considered the latent dimension $Q = 5$ and independently ran GPFA and OSLMM on the four datasets. The prediction error and coefficient of determination for the four datasets are reported in Figure~\ref{fig:prediction_ECoG}C and Figure~\ref{fig:prediction_ECoG}D. These results show that OSLMM provides robustly better predictive performance than GPFA on single-trial analysis. Moreover, we provided the analysis of the relation between the predictive performance and latent dimension size $Q$ in \nameref{S5_Appendix}. It shows that for most of combinations of the stimuli $\R{i}$ and latent dimension size $Q$, OSLMM outperforms GPFA in predictive performance. 

We also employed GPFA and OSLMM on all 4800 single-trial data and conducted global analysis. The leave-one-channel-out prediction with a three-fold cross validation is conducted for model comparison. In the experiment, we initialized the latent variables in OSLMM with the estimate from GPFA as good starting points. For each stimulus, we reported the prediction error (sum of square error) in the logarithmic scale (smaller is better) and reported the coefficient of determination $R^2$ (larger is better) in  Figure~\ref{fig:prediction_ECoG}E and Figure~\ref{fig:prediction_ECoG}F. The smaller log prediction error implies to the better prediction performance while the larger $R^2$ refers to better prediction performance. Figure~\ref{fig:prediction_ECoG}E and Figure~\ref{fig:prediction_ECoG}F visually show that log prediction errors in majority of channels from OSLMM are significantly smaller than those from GPFA and the opposite of behavior for $R^2$. It suggests that OSLMM provides better prediction performance. Quantitatively, we conducted the Wilcoxon sign-ranked test on the log prediction error and $R^2$, and the $p$ values are $1.16 \times 10^{-38}$ and $4.56 \times 10^{-38}$ respectively. It demonstrated that OSLMM model significantly outperforms GPFA in model prediction.

\subsection*{OSLMM extracts interpretable representations that reflect the distribution of stimulus}

\begin{figure}[ht!]
    \centering
    \includegraphics[width=\textwidth]{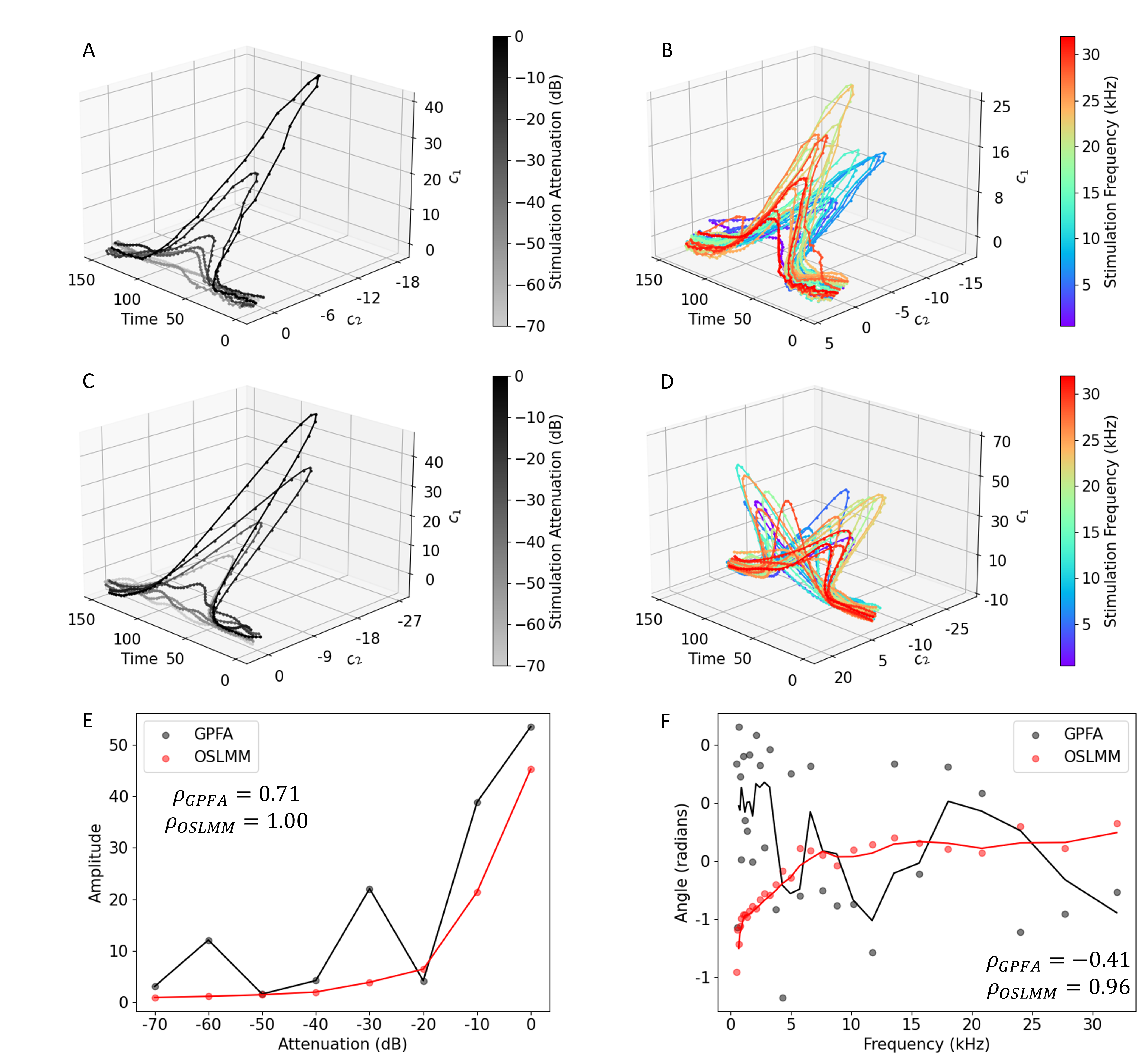}
    \caption{OSLMM extracts interpretable representations that reflect the distribution of stimulus: Inferred orthonormalized latent functions from OSLMM and GPFA for all stimuli.
    (A, C) Eight stimuli for all attenuation with a fixed frequency $\SI{7627}{\hertz}$ averaged by trials. (A) OSLMM; (C) GPFA);  (B, D): The same type of inferred orthonormalized latent functions for OSLMM (B) and GPFA (D) but for all frequencies with a fixed attenuation $-10$ dB averaged by trials. (E, F): Summary plots for GPFA and OSLMM. We plotted the attenuation vs and amplitude of the latent trajectory in (E) and plotted the frequency vs the angle of latent trial in (F). We provided the Spearman's correlations for GPFA and OSLMM in (E) and (F). Moreover, we conducted linear regression between the peak of latent trajectories and exogenous variable (attenuation or frequency).
    The $R^2$ scores for OSLMM/GPFA are $0.84/0.62$(Frequency: $7627$ Hz) and $0.50/0.06$(Attenuation: $-10$ dB)}.
    \label{fig:lr_ECOG}
\end{figure}

We applied both GPFA and OSLMM to jointly model the trials of all different stimuli, and explored the structure of the latent functions. For both methods, we set the latent dimension $Q=5$, and then inferred the latent functions of all trials. Specifically, we estimated the shared model parameters $\B{S}$ and individual latent functions $\B{f}$ using their corresponding posterior mean. We converted individual latent functions $\B{f}$ to the individual orthonormalized latent functions $\B{c}$ with the estimate $\B{S}$. Latent functions are rotated to maximize the power captured by each latent in decreasing order. Finally, we averaged the orthonormalized latent functions by stimuli over its corresponding trials and plot them in Figure~\ref{fig:lr_ECOG}A and Figure~\ref{fig:lr_ECOG}C. For comparison, we plotted the averaged orthonormalized neural trajectories for the stimuli in GPFA in Figure~\ref{fig:lr_ECOG}B and Figure~\ref{fig:lr_ECOG}D. 

In particular, we plotted the averaged orthonormalized latent functions for all eight stimuli with a fixed frequency $\SI{7626}{\hertz}$ in Figure~\ref{fig:lr_ECOG}A and Figure~\ref{fig:lr_ECOG}C, and we plotted the averaged orthonormalized latent functions for all thirty frequencies with a fixed attenuation $-10$ dB in Figure~\ref{fig:lr_ECOG}B and Figure~\ref{fig:lr_ECOG}D. We found that OSLMM latent functions accurately reflected the monotonic ordering of both the different amplitudes (Figure~\ref{fig:lr_ECOG}A) and frequencies (Figure~\ref{fig:lr_ECOG}B). In contrast, GPFA latent dimensions did not have this property (Figure~\ref{fig:lr_ECOG}C and Figure~\ref{fig:lr_ECOG}D). Specifically, the trajectories for different amplitudes extended in the direction of the two OSLMM latent functions (Figure~\ref{fig:lr_ECOG}A) with a magnitude that increased monotonically with increasing sound amplitude (grey-to-black), while the GPFA trajectories had mixed ordering (Figure~\ref{fig:lr_ECOG}C). Likewise, trajectories for different sound frequencies (blue-to-red) smoothly transitioned across the first OSLMM latent function (Figure~\ref{fig:lr_ECOG}B), but were highly intermixed in the GPFA trajectories (Figure~\ref{fig:lr_ECOG}D). Thus, the OSLMM trajectories reflect expected distributed auditory cortical population response properties for both of these stimulus dimensions. We plotted the attenuation vs the amplitude of latent trajectory in Figure~\ref{fig:lr_ECOG}E and plotted the frequency vs the angle of latent trajectory in Figure~\ref{fig:lr_ECOG}F. Both plots show that the monotonous property between attenuation and amplitude and the monotonous property between frequency and angle in OSLMM. But there exists no monotonous properties in GPFA. It suggests that OSLMM provides more interpretable latent trajectories than those from GPFA. Quantitatively, we provided the Spearman correlation for GPFA and OSLMM between attenuation and amplitude, and between frequency and angle in Figure~\ref{fig:lr_ECOG}E and Figure~\ref{fig:lr_ECOG}F. Both results illustrate the stronger monotonous property in OSLMM than that in GPFA.  
Moreover,we conducted linear regression between the peak of latent trajectories and exogenous variables. The $R^2$ scores for OSLMM/GPFA are 0.84/0.62(Frequency: 7627 Hz) and 0.50/0.06(Attenuation: -10 dB). It suggests that the latent representation inferred from OSLMM is more informative than that inferred from GPFA to decode the exogenous variables (Stimulation attenuation and stimulation frequency). 

On the other hand, we provided more visualization results under different latent dimension sizes $Q$ in \nameref{S6_Appendix}, which shows that the primary principle components are relatively robust to the selection of $Q$.

\subsection*{Analysis on monkey arm-reaching experiments}
We applied GPFA and OSLMM for global analysis on monkey arm-reaching experiments. 
We first showed the extracted representations from OSLMM are more predictive to behavior than those from GPFA in terms of monkey's hand positions.
Then we showed that the extracted representations from OSLMM are more interpretable than those from GPFA in the sense of delivering trajectories which better reflect expected reach angle distributions as well as velocity distributions. In the following experiments, we set thte latent dimension size $Q=6$.

\subsubsection*{OSLMM's latent representations are predictive to the monkey's behavior}

\begin{figure}[ht!]
    \centering
    \includegraphics[width=\textwidth]{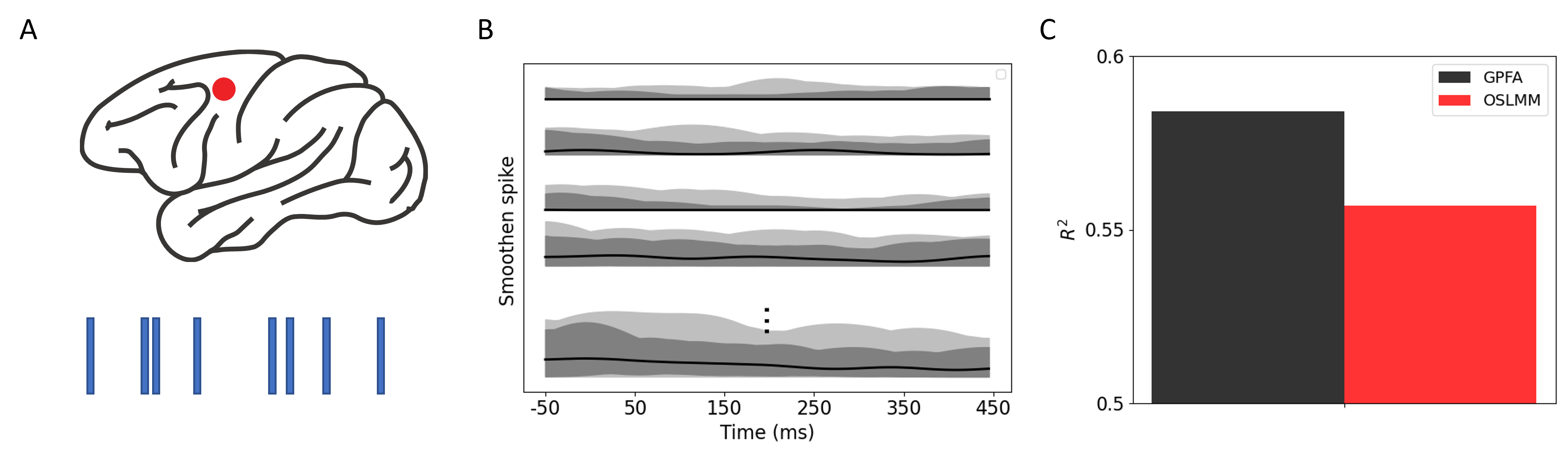}
    \caption{OSLMM's latent representations are predictive to the monkey's behavior: Spike data and decoding performance on the monkey's hand positions. We extracted the neural spike data from monkeys' motor cortex (A) and smoothen spikes via Gaussian convolution with 50ms band. We provided the functional boxplots of the smoothen spikes where black lines refer to the median curve and light/dark grey shaded areas refer to functional data envelopes and $50\%$ central region. (C) Coefficient of determination $R^2$ score on the decoding tasks for GPFA and OSLMM.}
    \label{fig:decoding_MAZE}
\end{figure}

We first preprocessed the spike data (Figure~\ref{fig:decoding_MAZE}A) via the Gaussian convolution with 50ms band. We visualized the smoothen spikes in each channel across my trials in Figure~\ref{fig:decoding_MAZE}B using the functional boxplots \cite{sun2011functional,meng2021nonstationary} where black lines refer to the median curve and light/dark grey shaded areas refer to functional data envelopes and $50\%$ central region. 

We decoded the monkey's hand position solely from the inferred latent functions from GPFA or OSLMM via ridge regression where the hyper-parameters are selected through cross validation. The decoding R2 scores of GPFA and OSLMM are $0.557$ and $0.584$ respectively (Figure~\ref{fig:decoding_MAZE}). It shows that OSLMM provides better predictive performance on behavior decoding.

\subsubsection*{OSLMM extracts interpretable representations that reflect the distribution of reach angle and velocity}

\begin{figure}[ht!]
    \centering
    \includegraphics[width=\textwidth]{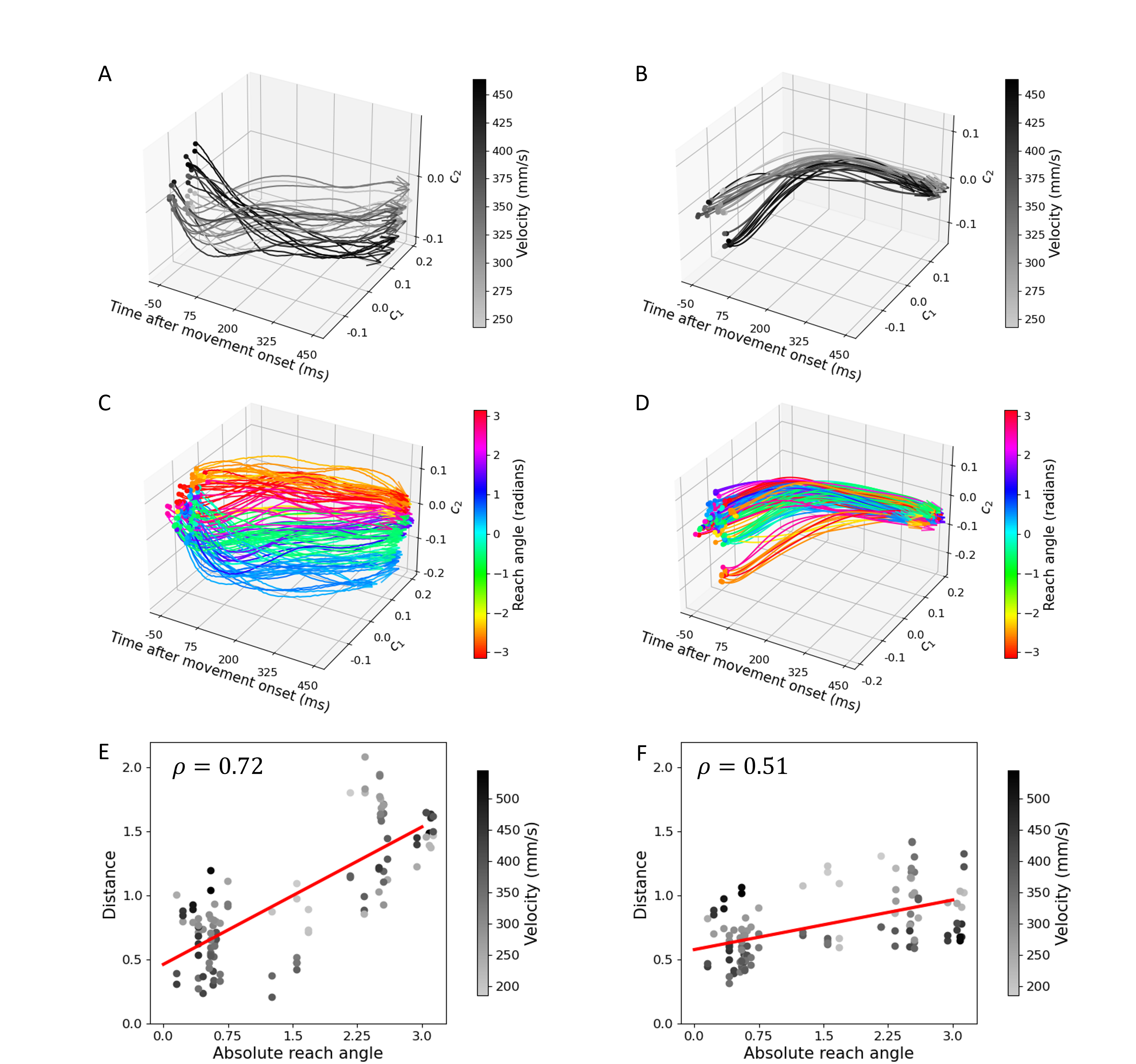}
    \caption{OSLMM extracts interpretable representations that reflect the distribution of reach angle and velocity: Inferred orthonormalized latent functions from OSLMM and GPFA with colors encoded by average speeds and reach angles. (A-B) We take all conditions whose reach angles are in (-1, 0) radians and encode the average speeds into colors. (A) OSLMM; (B) GPFA);  (C-D): The same type of inferred orthonormalized latent functions for OSLMM(C) and GPFA (D) but for all conditions with colors encoded by reach angles. (E-F) We averaged the latent trajectories whose reach angle is within (-0.5, 0.5) radians as baseline trajectory and computed the $\ell_2$ distance between individual latent trajectory and baseline trajectory as well as the absolute reach angle. We reported the spearman correlations between absolute reach angle and distance for both methods in (E-F). Moreover, we visualized the relation between absolute reach angle and distance using a scatterplot with colors encoded by their velocities and fitted a linear model between the absolute reach angles and $\ell_2$ distance and the p-values of the linear coefficient for OSLMM (E) and GPFA (F) are $1.25\times 10^{-23}$ and $4.17\times 10^{-9}$.}
    \label{fig:latent_maze}
\end{figure}

We applied GPFA and OSLMM for all trials of smoothed spike data. Then we estimated the orthogonalized latent functions and rotated them to maximize the power captured by each latent in decreasing order. We averaged those latent functions grouped by conditions and visualized the first two latent principle components vs time in Figure~\ref{fig:latent_maze}. Particularly, we selected conditions whose reach angles are in (-1, 0) radians and encoded the average velocity into colors on the orthogonalized latent functions from OSLMM and GPFA, shown in  Figure~\ref{fig:latent_maze}A and Figure~\ref{fig:latent_maze}B. We also encoded the reach angles for all conditions on the orthogonalized latent functions from OSLMM and GPFA, shown in Figure~\ref{fig:latent_maze}C and Figure~\ref{fig:latent_maze}D. 

The latent trajectories of OSLMM can better reflect the change of the velocity than GPFA. In general, the latent trajectories with less velocity have an increasing trend while those with higher velocity have a decreasing trend on the second principle component for OSLMM in Figure~\ref{fig:latent_maze}A. For GPFA in Figure~\ref{fig:latent_maze}B, although there are some difference of latent trajectories between the conditions with low and high velocities in the first half time interval time, all latent trajectories mixed with each other in the last half time interval.

As for the relation between latent trajectories and reach angles, we find that the reach angle information can be easily decoded from the latent trajectories in OSLMM. In particular, for the second principle component, as the value varies from low to high, it directly matches to the color from blue to orange via green or red. In the other words, as the value in the second principle component increase, the reach angles are deviating the zero radians. But for GPFA, all colors are mixed as time increases and it is hard to find the relation between latent trajectories and reach angles (colors). Therefore, it illustrates that OSLMM provides more interpretable and separable pattern in latent space than GPFA according to different reach angles.

In addition, to quantify the performance of the decoding behavior from latent representation to exogenous variable (reach angle), we conducted another quantitative analysis. The underlying motivation of this analysis is that the better decoding behavior the model has, the more similar the topology of latent represent and exogeneous variable is. We measure the topology by the distance between the baseline and individual. In the other words, the more far away from the zero radians the reach angle is, the far away from the baseline trajectory the corresponding latent trajectory locates, where the baseline trajectory matches the zero reach angle.

Specifically, we first averaged the latent trajectories whose reach angle is within (-0.5, 0.5) radians and defined it as baseline trajectory. Then we computed the $\ell_2$ distance between individual latent trajectory and baseline trajectory paired with the corresponding absolute reach angle. We draw the scatter plot for those pairs with colors encoded by velocity. 

Next, we reported the spearman correlations between distance and absolute reach angle for GPFA and OSLMM as $\rho_{GFPA} = 0.51$ and $\rho_{OSLMM} = 0.72$. It implies that the topology of latent trajectories in OSLMM is more similar to that of reach angle compared with latent trajectories in GPFA. Alternative, we fitted a linear model between the absolute reach angles and $\ell_2$ distance for OSLMM (Figure~\ref{fig:latent_maze}E) and GPFA (Figure~\ref{fig:latent_maze}F) and reported the p-value of the linear coefficient. The p-value for OSLMM is $1.25 \times 10^{-23}$ while the p-value for GPFA is $4.17 \times 10^{-9}$. The p-value for OSLMM is significantly smaller than that for GPFA. It validates the conclsion that comparison with GPFA, the topology of latent trajectories from OSLMM is more similar to that of reach angles. As for the relation between latent trajectories and velocities, we found no significant dependence for both methods in Figure~\ref{fig:latent_maze}E and Figure~\ref{fig:latent_maze}F. But we note that the dependence would be easily visualized for OSLMM in Figure~\ref{fig:latent_maze}A when the whole representations instead of the distance are visualized. 

Furthermore, to robustly illustrate that the latent trajectories from OSLMM have more similar topology to reach angles than those from GPFA, we conducted bootstrap method on the p-values in the above analysis and carried out the Wilcoxon sign-ranked test on the pair-wise difference between GPFA and OSLMM. We chose 100 bootstrap samples and the full sample size. The p-value of the hypothesis test is $3.90 \times 10^{-18}$, implying the OSLMM performs significantly better than GPFA in the sense of better reflecting the order of reach angles. 

\begin{figure}[ht!]
    \centering
    \includegraphics[width=\textwidth]{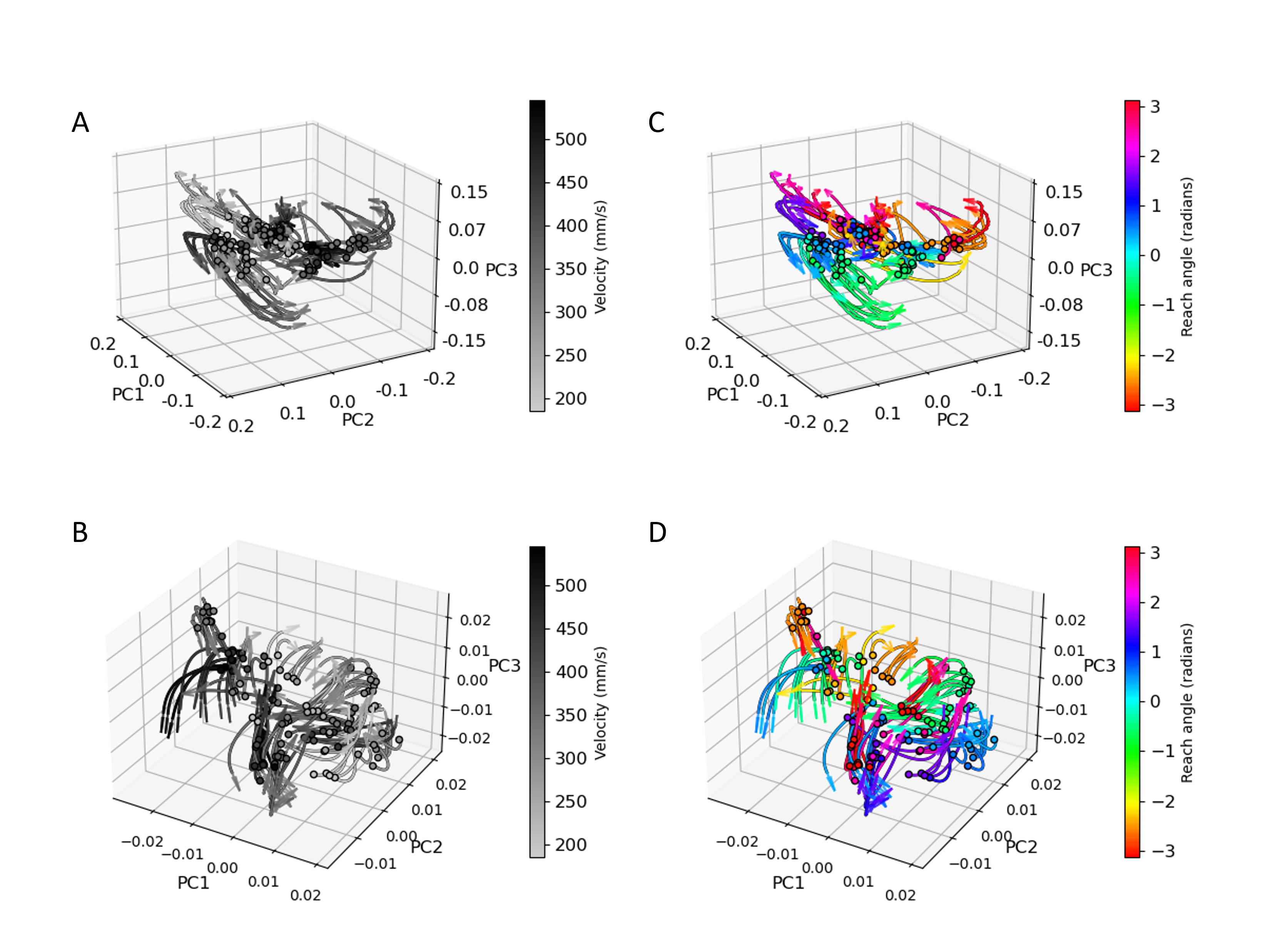}
    \caption{Inferred representations of the first three principle components of jPCA from OSLMM (A, C) and GPFA (B, D). We encoded the velocities into colors in (A, B) and encoded the reach angles into colors in (C, D).}
    \label{fig:latent_maze_jpca}
\end{figure}

To additionally study the structure of latent trajectories, we conducted the jPCA analysis \cite{churchland2012neural} based on the extracted latent representations from GPFA and OSLMM. We visualized the first three jPCs with first $30$ time stamps in Figure~\ref{fig:latent_maze_jpca}. Latent representations from OSLMM are displayed in Figure~\ref{fig:latent_maze_jpca}A and Figure~\ref{fig:latent_maze_jpca}C and those from GPFA are displayed in Figure~\ref{fig:latent_maze_jpca}B and Figure~\ref{fig:latent_maze_jpca}D. We encoded velocities into colors in Figure~\ref{fig:latent_maze_jpca}A and Figure~\ref{fig:latent_maze_jpca}B and encoded reach angles into colors in Figure~\ref{fig:latent_maze_jpca}C and Figure~\ref{fig:latent_maze_jpca}D. Comparing Figure~\ref{fig:latent_maze_jpca}A with Figure~\ref{fig:latent_maze_jpca}B, it shows that the representations from OSLMM have stronger dependence with velocity than those from GPFA. It is because that the representations with small velocities are clustered on the upper left corner with clockwise direction in Figure~\ref{fig:latent_maze_jpca}A while those representations in Figure~\ref{fig:latent_maze_jpca}B are randomly distributed on the upper right in Figure~\ref{fig:latent_maze_jpca}B. On the other hand, as for the relation between representations and reach angles in Figure~\ref{fig:latent_maze_jpca}C and Figure~\ref{fig:latent_maze_jpca}D. Through the anticlockwise direction, the color in Figure~\ref{fig:latent_maze_jpca}C vary from green to yellow, red and blue which matches the decreasing order (looped) in reach angles. But no clear dependence between representations and reach angles exists in Figure~\ref{fig:latent_maze_jpca}D. Hence, it illustrates that OSLMM would contributes to more interpretable representations in the jPCA analysis.

%% file: DIS.tex
\section*{Discussion} \label{sec:discussion}

We have proposed a new multi-output regression framework, the orthogonal stochastic linear mixing model (OSLMM). Our proposed model can capture input-dependent correlations across outputs and enable accurate prediction by utilizing an adaptive mixing mechanism, where mixing coefficients depend on inputs. We note that, like GPRN, OSLMM is strictly a non-Gaussian model due to its adaptive mixing mechanism. Moreover, by imposing an orthogonal constraints on the coefficient matrices, MCMC inference scales linearly with the output dimension $P$ and the number of latent functions $Q$, allowing efficient scaling to large datasets. This is achieved by breaking down the high dimensional prediction problem into independent single-output problems to sample latent functions and using efficient MCMC to sample the orthogonal space on the Steifel manifold. Together, these features enable the method to analyze large datasets with complicated input-dependent correlations across many outputs. We demonstrated the numerical superiority of OSLMM in various real-world benchmark datasets. Finally, we used OSLMM for single-trial analysis of neural data, demonstrating that it provides not only better prediction performance but also more interpretable latent representations than GPFA. Together, these results indicate that OSLMM will be beneficial for analysis of many high-dimensional timeseries datasets, especially for neural data.

We applied OSLMM in two neurosicence datasets. One is rat auditory cortex dataset and the other is monkey arm-reaching dataset. We find the extracted latent representations from OSLMM can be easily used to decode the exogenous variables. In the other words, the latent representations in OSLMM can reflect the data generation process, which would benefit us to better understand the data generation process in neural responses and would potentially provide the same benefits understand the data generation process in other neuroscience experiments.

A limitation of OSLMM is that, when the number of samples is very large, sampling all latent functions is still expensive. As a potential future work, variational inference for OSLMM may overcome this issue.

%% file: SUP.tex
\paragraph*{S1 Appendix}
\label{S1_Appendix}
\subsection*{Hyper-parameter learning for SLMM}

When considering the independent noise such that $\Sigma$ is a diagonal matrix, we set the a conjugate inverse Gamma prior $p(\Sigma) = \prod_{p=1}^P \mathcal{IG}(\sigma^2_p| a, b)$, where $\sigma^2_p$ is the $p$th element on the diagonal of $\Sigma$. Then the conditional posterior distribution of $\sigma_p^2$ is
\begin{align}
    \sigma_p^2| - & \propto \prod_{t=1}^T \mathcal{N}(\R{y}_{tp}| \R{g}_{tp}, \sigma_p^2)\mathcal{IG}(\sigma^2_p| a, b) \nonumber \\
    & \sim \mathcal{IG}(\sigma^2_p| a + \frac{N}{2}, b + \frac{\sum_{n=1}^N(\R{y}_{tp} - \R{g}_{tp})^2}{2}) \,. \label{eq:pos_sigma2_p_slmm}
\end{align}
In practice, we set $a = 0.01$ and $b=0.01$ to allow large variance.

We consider the commonly-used squared exponential (SE) covariance function for $W$ and $f$
\begin{align}
    K_i(\R{t}_1, \R{t}_2) = \sigma^2_{i} \exp(-\frac{\|\R{t}_1 - \R{t}_2\|^2}{2 l_i^2})
\end{align}
where $i = W$ or $f$. $\sigma^2_f = 1$ is fixed for model identifiability. We put a conjugate prior on $\sigma^2_W$ such that $\sigma_W^2 \sim \mathcal{IG}(c, d)$. Then the conditional posterior distribution is
\begin{align}
    \sigma_W^2|- & \propto \prod_{p=1}^P\prod_{q=1}^Q \mathcal{N}(\B{w}_{pq}| \bm 0, \sigma^2_W \tilde{\B{K}}_w) \mathcal{IG}(\sigma^2_W| c, d) \nonumber \\
    & \sim \mathcal{IG}(\sigma^2_W| c + \frac{NPQ}{2}, d + \frac{\sum_{i=1}^P\sum_{j=1}^Q \B{w}_{pq}' \tilde{\B{K}}_W^{-1} \B{w}_{pq}}{2})
\end{align}
where $\tilde{\B{K}}_W$ is the correlation matrix and $\B{K}_w = \sigma^2_W \tilde{\B{K}}_w$. As for length-scale parameters $l_i^2$, we put a non-informative prior $l_i^2 \propto \frac{1}{l_i^2}$ and sample them via adaptive Metropolis-with-Gibbs algorithm \cite{roberts2009examples}.

\paragraph*{S2 Appendix}
\label{S2_Appendix}
\subsection*{Theoretical proofs for sufficient statistics}

\textbf{Theorem} $\B{T}_n\B{y}_n$ is a minimally sufficient statistic for $\B{f}_n$.

\textbf{Proof}: 
Without loss of generality, we ignore the subscript $n$ in this proof. To show $\B{T}\B{y}$ is a minimally sufficient statistic for $\B{f}$, we need to prove $p(\B{y}_1|\B{f}) / p(\B{y}_2|\B{f})$ is a constant as a function of $\B{f}$ if and only if $\B{T}\B{y}_1 = \B{T}\B{y}_2$. We have 
\begin{align}
    \log\frac{p(\B{y}_1|\B{f}) }{p(\B{y}_2|\B{f}) } & = \log\frac{\mathcal{N}(\B{y}_1| \B{U}\B{S}^{\frac{1}{2}} \B{f}, \Sigma)}{\mathcal{N}(\B{y}_2| \B{U}\B{S}^{\frac{1}{2}} \B{f}, \Sigma)} \nonumber \\
    & =  (\B{y}_1 - \B{y}_2)'\Sigma^{-1} \B{U}\B{S}^{\frac{1}{2}} \B{f} + \R{const}\nonumber \\
    & = \B{f}' \B{S}^{\frac{1}{2}} \B{U}'\Sigma^{-1} (\B{y}_1 - \B{y}_2) + \R{const}\nonumber
\end{align}
When we consider the homogeneous noise $\Sigma = \sigma_y^2\B{I}$, we have
\begin{align}
    \log\frac{p(\B{y}_1|\B{f}) }{p(\B{y}_2|\B{f}) } & = \frac{1}{\sigma^2_y} \B{f}' \B{S}^{\frac{1}{2}} \B{U}'(\B{y}_1 - \B{y}_2) + \R{const}\nonumber \\
    & = \frac{1}{\sigma^2_y} \B{f}' \B{S}^{\frac{1}{2}} \B{U}'\B{U}\B{S}^{\frac{1}{2}}\B{S}^{-\frac{1}{2}}\B{U}'(\B{y}_1 - \B{y}_2) + \R{const}\nonumber \\
    & = \B{f}' \B{S}^{\frac{1}{2}} \B{U}'\Sigma^{-1}\B{U}\B{S}^{\frac{1}{2}}\B{T}(\B{y}_1 - \B{y}_2) + \R{const} \,. \label{eq:ratio_prob}
\end{align}

Because $\B{S}^{\frac{1}{2}} \B{U}'\Sigma^{-1}\B{U}\B{S}^{\frac{1}{2}}$ is invertible, Equation~\ref{eq:ratio_prob} does not depend on $\B{f}$ if and only if $\B{T}\B{y}_1 = \B{T}\B{y}_2$. Therefore, $\B{T}_n\B{y}_n$ is a minimally sufficient statistic for $\B{f}_n$.

\paragraph*{S3 Appendix}
\label{S3_Appendix}
\subsection*{Hyper-parameter learning for OSLMM}
We consider the homogeneous noise such that $\Sigma = \sigma^2_y \B{I}$ in this setting and we put a conjugate prior on the variance, $p(\sigma_y^2) =  \mathcal{IG}(\sigma^2| a, b)$. The conditional posterior distribution is 
\begin{align}
    \sigma_y^2| - & \propto \prod_{t=1}^T\mathcal{N}(\B{y}_{t}| \B{g}_{t}, \sigma_y^2\B{I}) \mathcal{IG}(\sigma_y^2|a, b) \nonumber \\
    & \sim \mathcal{IG}(\sigma_y^2 | a + \frac{PT}{2}, b + \frac{\sum_{t=1}^T(\R{y}_{td} - \R{g}_{td})^2}{2}) \,.
\end{align}

We consider the commonly-used SE covariance function for $\bm h$ and $\bm f$. $\sigma^2_f = 1$ is fixed for model identifiability. We put a conjugate prior on $\sigma_h^2$ such that $\sigma_h^2 \sim \mathcal{IG}(c, d)$. Then the conditional posterior distribution is 
\begin{align}
    \sigma_h^2|- & \propto \prod_{q=1}^Q \mathcal{N}(\B{h}_{q}| \bm 0, \sigma^2_h \tilde{\B{K}}_h) \mathcal{IG}(\sigma^2_h| c, d) \nonumber \\
    & \sim \mathcal{IG}(\sigma^2_W| c + \frac{QT}{2}, d + \frac{\sum_{q=1}^Q \B{h}_{q}' \tilde{\B{K}}_h^{-1} \B{h}_{q}}{2})
\end{align}
where  $\tilde{\B{K}}_h$ is the correlation matrix and $\B{K}_h = \sigma^2_h \tilde{\B{K}}_h$.

\paragraph*{S4 Appendix}
\label{S4_Appendix}
\subsection*{Prediction comparison on real datasets}
We compared SLMM and OSLMM to GPRN models with the following inference approaches: (1) MFVB -- mean-field variational Bayes inference \cite{wilson2011gaussian}, (2) NPV -- nonparametric variational Bayes inference \cite{nguyen2013efficient}, (3)SGPRN -- scalable variational Bayesian inference \cite{li2020scalable}. For both SLMM and OSLMM, Markov Chain Monte Carlo had $500$ iterations, in which the first $200$ iterations are used for burnin. For the variational methods, GPRN(MFVB) and GPRN(NPV) ran $100$ iterations and SGPRN ran $2000$ epochs to ensure convergence. 

We evaluated the model performances on five real-world datasets, \textbf{Jura}, \textbf{Concrete}, \textbf{Equity}, \textbf{PM2.5} and \textbf{Neural}, with $3, 3, 25, 100$ and $128$ outputs respectively. Specifically, (1) \textbf{Jura}, the concentrations of cadmium at $100$ locations within a $14.5$ $\R{km}^2$ region in Swiss Jura. Following \cite{li2020scalable}, we utilized the concentrations of cadmium, nickel, and zinc at $259$ nearby locations to predict the three correlated concentrations at another $100$ locations. (2) \textbf{Concrete}, a geostatistics dataset, including 103 samples with 7 concrete mixing ingredients as input variables and with 3 output variables (slump, flow, and compressive strength). We random split it into a training set of $80$ points and a test set of $23$ points as in \cite{nguyen2013efficient}. (3)\textbf{Equity}, a financial dataset consists of 643 records of 5 equity indices. The task is to predict the $25$ pairwise correlations. Following \cite{wilson2011gaussian} we randomly chose 200 records for training and chose another 200 records for testing.  (4) \textbf{PM2.5}, 100 spatial measurements of the particulate mater pollution (PM2.5) in Salt Lake City in July 4-7, 2018, where inputs are time stamps. We randomly took 256 samples for training and 32 for testing. (5) \textbf{Neural}, a micro-electrocorticography (\SI{}{\micro\metre} ECoG) recordings from rat auditory cortex in response to pure tone pips collected in the Bouchard Lab \cite{dougherty2019laminar}.We randomly selected $100$ samples for training and another $100$ for testing. For all datasets, we normalized each input dimension to have zero mean and unit variance; for \textbf{Jura}, \textbf{Concrete} and \textbf{Neural} data, the outputs in each dimension are normalized to have zero mean and unit variance.

We report the predictive mean absolute error for datasets with moderate-to-large output dimension \textbf{Equilty}, \textbf{PM2.5} and \textbf{Neural} in Table~\ref{tab:prediction_real_data}. For datasets with small output dimension (\textbf{Jura} and \textbf{Concrete}), the predictive performance of OSLMM does not significantly outperform other methods, and gives similar results to GPRN(NPV). This may be because the output correlation is trivial. We provide the predictive mean absolute error for those two datasets in Appendix A. All results were summarized by the mean and standard deviation over 5 runs. Table~\ref{tab:prediction_real_data} shows that the prediction performance of OSLMM is uniformly and robustly better than the other four methods.

\begin{table}[ht!]
    \centering
    \caption{Predictive mean absolution error of five methods on three real datasets, \textbf{Equilty}, \textbf{PM2.5} and \textbf{Neural}. The results were summarized by mean and standard deviation over 5 runs.}
    \begin{tabular}{c|c|c|c}
         \hline
                & \textbf{Equity} & \textbf{PM2.5} & \textbf{Neural} \\ 
         \hline
         SLMM   & 2.6995e-5 (7.6614e-7) & 9.5514 (0.3703) & 0.6068 (0.0018) \\
         OSLMM & \textbf{2.6643e-5} (2.5686e-7) & \textbf{3.9699} (0.2595) & \textbf{0.5141} (0.0206) \\
         GPRN (MFVB) & 3.0327e-5 (8.1183e-7) & 5.9738  (1.3893) & 0.5654 (0.0047) \\
         GPRN (NPV) & 4.3490e-5 (5.9300e-6)  & 6.1794 (1.4397) & 0.5724 (0.0051)\\
         SGPRN & 2.7346e-5 (1.4374e-7) & 8.6163 (2.1070) & 0.5727 (0.0263)\\
         \hline
    \end{tabular}
    \label{tab:prediction_real_data}
\end{table}

\begin{figure}
    \centering
    \includegraphics[width=\linewidth]{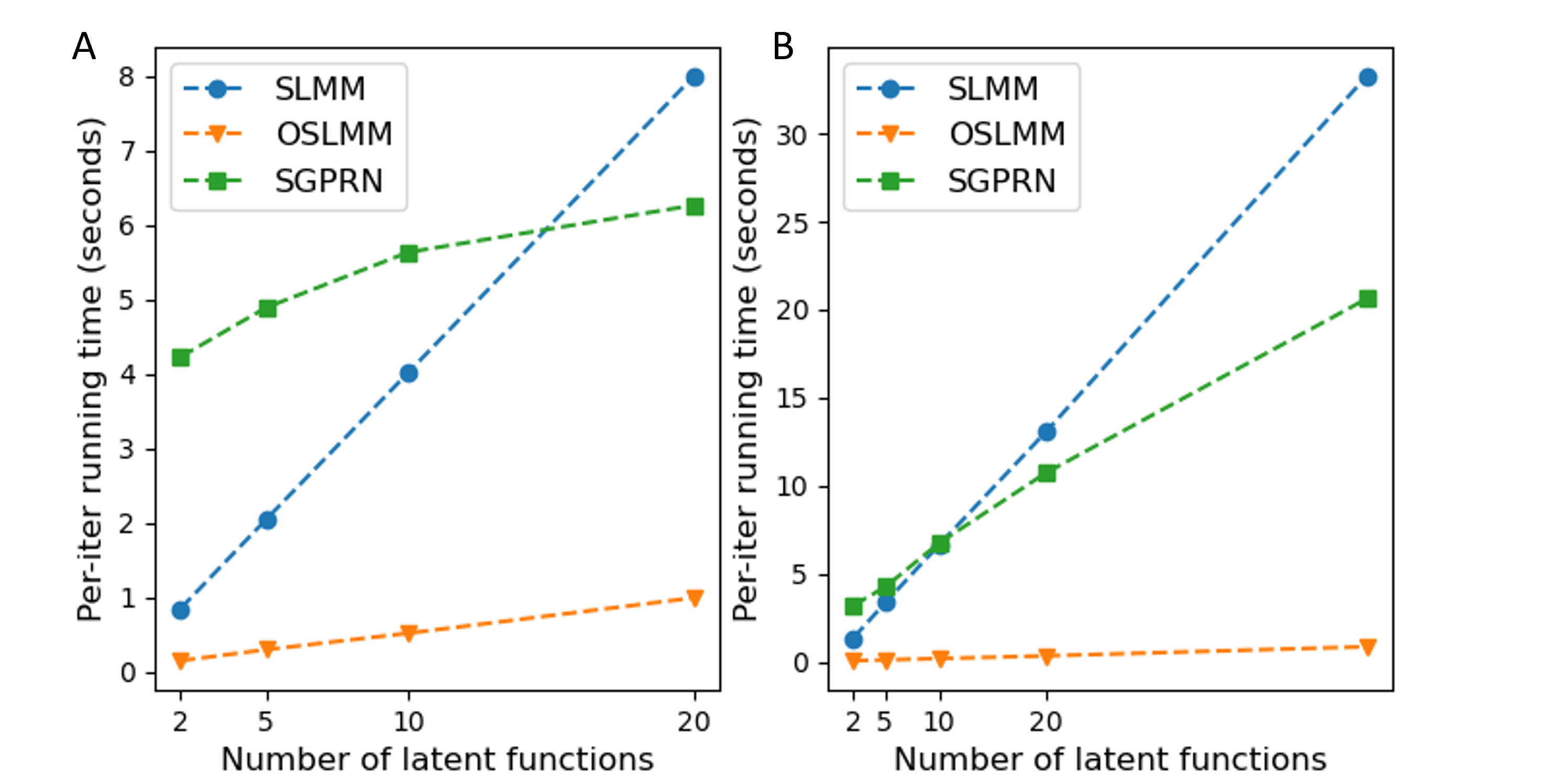}
    \caption{Training speed of SLMM, OSLMM and SGPRN inference algorithms on \textbf{Equity} data (A) and \textbf{PM2.5} data (B). We show the running time per iteration in the setting with different number of latent functions.}
    \label{fig:running time for others}
\end{figure}

Next, we compared SLMM, OSLMM and SGPRN in terms of compute speed, since GPRN(MFVB) and GPRN(NPV) are known to be very slow \cite{li2020scalable}. We report the per-iteration running time of SLMM and OSLMM, and the average time of $4$ epochs of SGPRN for a fair comparison. For all three methods, because the number of latent functions $Q$ should be smaller than output dimension, $Q < P$, we varied the size of the latent functions, $Q = (2, 5, 10, 20, 50)$ for \textbf{PM2.5} and \textbf{Neural} and the size $Q = (2,5,10,20)$ for \textbf{Equity}. We report the result of  in Figure~\ref{fig:running time} and the results of \textbf{Equity} and \textbf{PM2.5} in Figure~\ref{fig:running time for others}.These results clearly demonstrate that inference of OSLMM faster than SLMM and SGPRN. 

On the other hand, we reported the predictive mean absolution error of five methods on two real datasets, \textbf{Jura} and \textbf{Concrete} in Table~\ref{tab:prediction_real_data2}

\begin{table}[ht!]
    \centering
    \caption{Predictive mean absolution error of five methods on three real datasets, \textbf{Jura} and \textbf{Concrete}. The results were summarized by mean and standard deviation over 5 runs.}
    \begin{tabular}{c|c|c}
         \hline
                & \textbf{Jura} & \textbf{Concrete} \\ 
         \hline
        SLMM & 0.6643 (0.0103) & 0.7627 (0.0507)\\
        OSLMM & 0.6230 (0.0079) & \textbf{0.5305} (0.0245)\\
        GPRN (MFVB) & 0.6346 (0.0047) & 0.7145 (0.1560)\\
        GPRN (NPV) & \textbf{0.6218} (0.0113) & 0.5567 (0.0225) \\
        SGPRN & 0.6762 (0.0669) & 0.8331 (0.0199) \\
         \hline
    \end{tabular}
    \label{tab:prediction_real_data2}
\end{table}

\paragraph*{S5 Appendix}
\label{S5_Appendix}
\subsection*{Analysis between predictive performance and latent dimension size in ECoG dataset}
\begin{figure}[ht!]
    \centering
    \includegraphics[width=\textwidth]{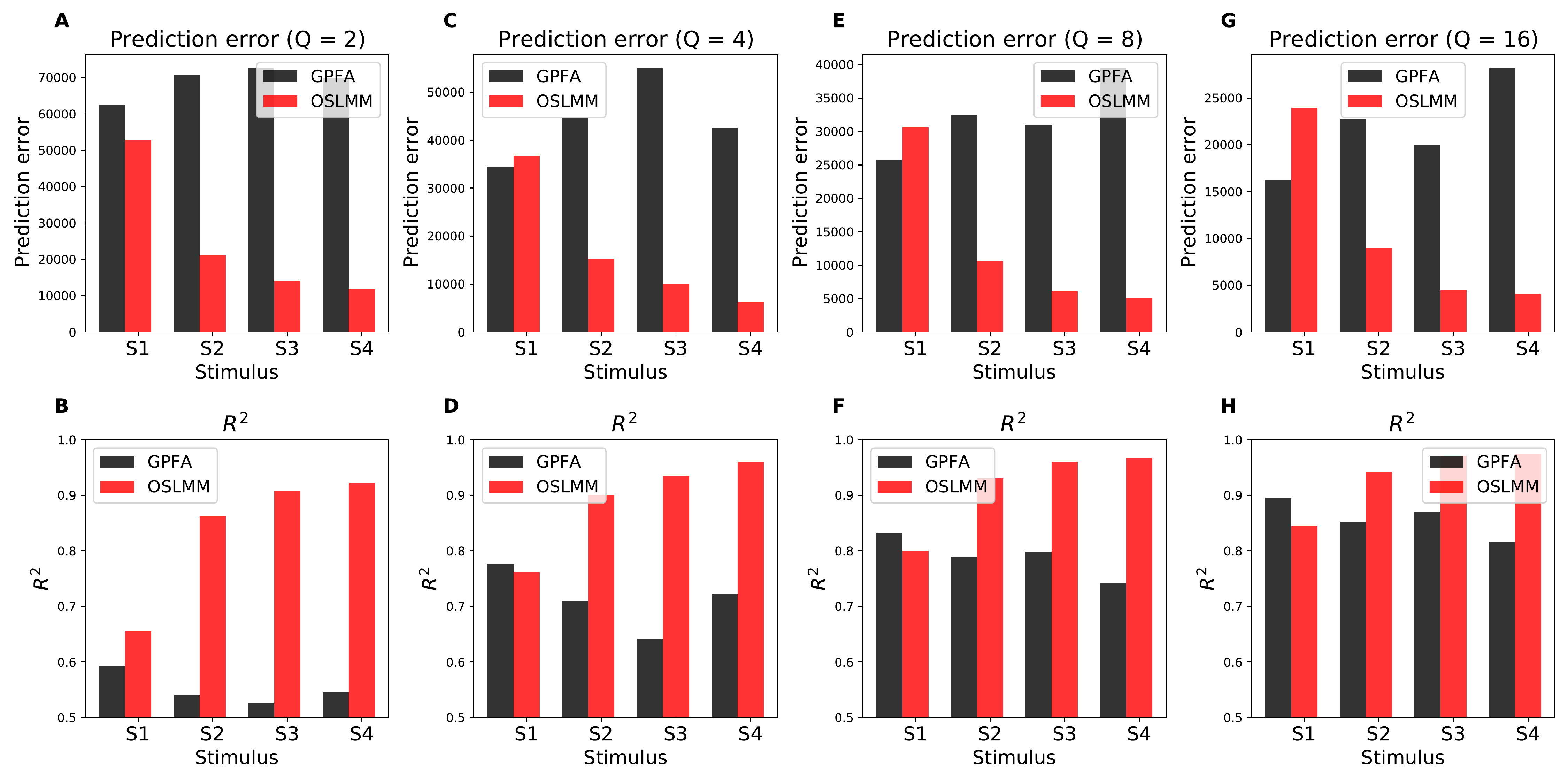}
    \caption{Prediction performance on leave-one-channel task on different latent dimension size $Q = 2, 4, 8$ and $16$}
    \label{fig:pred_vdims}
\end{figure}

We conduct leave-one-channel-out prediction tasks on the ECoG data for the same four stimuli S1, S2, S3 and S4 with different latent dimension $Q = 2, 4 ,8$ and $16$. We provide the prediction error and $R^2$ in Figure~\ref{fig:pred_vdims}. It shows that for most of channels and most of selection of $Q$, OSLMM outperforms GPFA in predictive performance. And we also find that when $Q > 2$, OSLMM outperforms GPFA for all four stimuli. 

\paragraph*{S6 Appendix}
\label{S6_Appendix}
\subsection*{Analysis between latent representation performance and latent dimension size in ECoG dataset}
\begin{figure}[ht!]
    \centering
    \includegraphics[width=\textwidth]{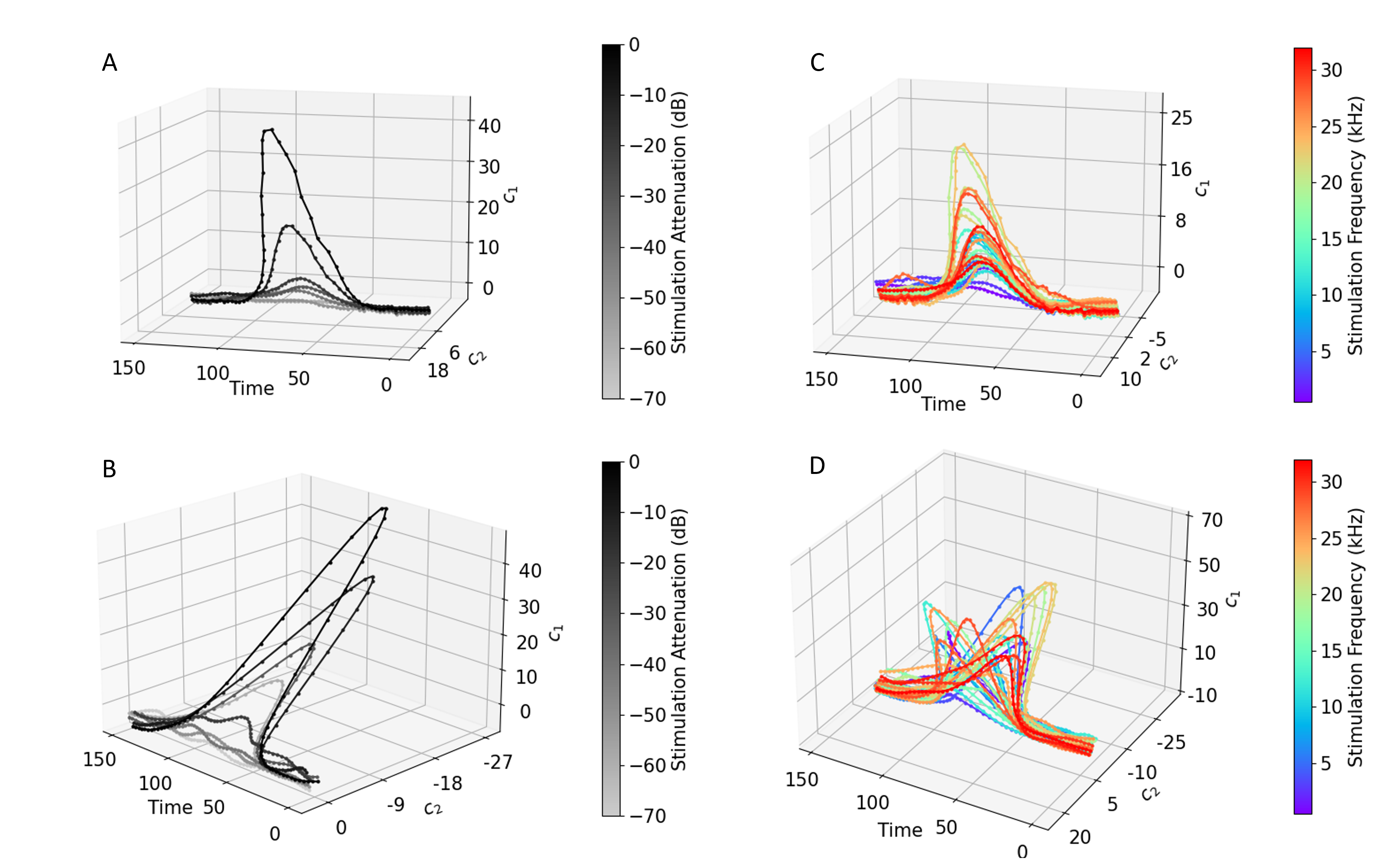}
    \caption{Inferred orthonormalized latent functions from OSLMM and GPFA for all stimuli with $Q = 10$.(A-B) Eight stimuli for all attenuation with a fixed frequency 7627 Hz averaged by trials. (A) OSLMM; (B) GPFA); (C-D):The same type of inferred orthonormalized latent functions for OSLMM (C) and GPFA (D) but for all frequencies with a fixed attenuation -10 dB averaged by trials. Moreover, we conducted linear regression between the peak of latent functions and exogenous variable (attenuation or frequency). The $R^2$ scores for OSLMM/GPFA are 0.71/0.61(Frequency: 7627) and 0.28/0.06(Attenuation: -10).}
    \label{fig:fig2_vdims10}
\end{figure}

\begin{figure}[ht!]
    \centering
    \includegraphics[width=\textwidth]{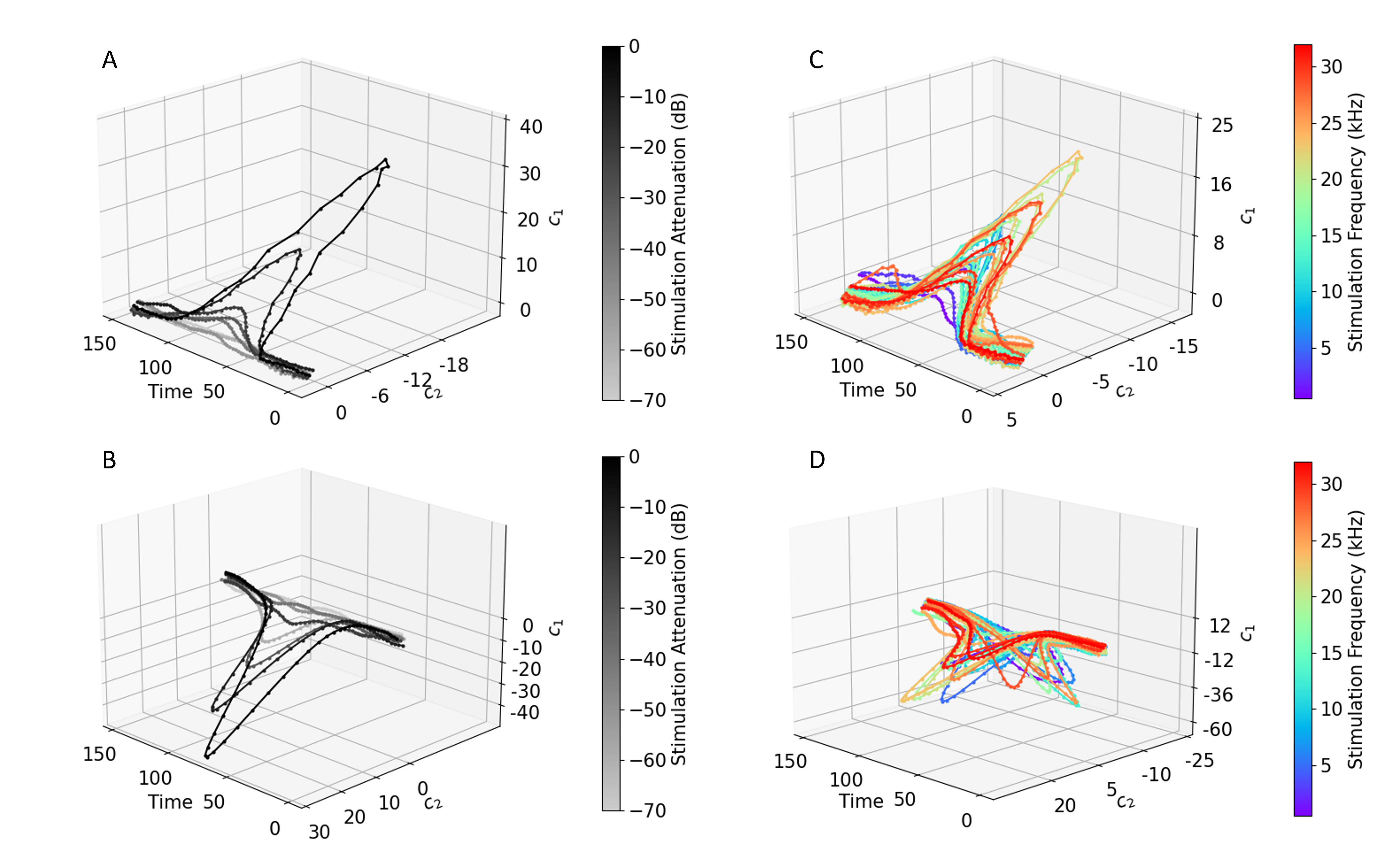}
    \caption{Inferred orthonormalized latent functions from OSLMM and GPFA for all stimuli with $Q = 15$.(A-B)Eight stimuli for all attenuation with a fixed frequency 7627 Hz averaged by trials. (A) OSLMM; (B) GPFA); (C-D):The same type of inferred orthonormalized latent functions for OSLMM (C) and GPFA (D) but for all frequencies with a fixed attenuation -10 dB averaged by trials. Moreover, we conducted linear regression between the peak of latent functions and exogenous variable (attenuation or frequency). The $R^2$ scores for OSLMM/GPFA are 0.85/0.62(Frequency: 7627) and 0.50/0.06(Attenuation: -10).}
    \label{fig:fig2_vdims15}
\end{figure}

We explore the relation between latent representation performance and latent dimension size by conducting OSLMM and GPFA on the ECoG data for all trials. We exploit different latent representation under different latent dimension size $Q = 5, 10$ and $15$. We display the first three principle components in the latent space in Figure~\ref{fig:lr_ECOG}, Figure~\ref{fig:fig2_vdims10} and Figure~\ref{fig:fig2_vdims15}. Those figures show that the latent representations of first three principle components have robust superior representations across different latent dimensions $Q$s.

